%% file: ms.tex
\documentclass[11pt]{article}
\pdfoutput=1
\usepackage[a4paper,bindingoffset=0.2in,%
            left=1in,right=1.2in,top=1in,bottom=1in,%
            footskip=.25in]{geometry}

\usepackage{amssymb}
\usepackage{amsmath}
\usepackage[english]{babel}
\usepackage{graphicx}
\usepackage{epstopdf}
\newtheorem{definition}{Definition}[section]
 
\newcommand{\citep}{\cite}
\newcommand{\citec}[2]{#1 \cite{#2}}

\begin{document}

\title{All-relevant feature selection using multidimensional filters with exhaustive search}

\author{Krzysztof Mnich k.mnich@uwb.edu.pl \\
       Computational Centre\\
       University of Bialystok\\
       Bialystok, Poland\\
       Witold R. Rudnicki w.rudnicki@uwb.edu.pl \\
       Institute of Informatics and Computational Centre\\
       University of Bialystok\\
       Bialystok, Poland\\
       Interdisciplinary Centre for Mathematical and Computational Modelling\\
       University of Warsaw\\
       Warsaw, Poland
       }



\maketitle

\input{fs0_abstract}

\input{fs1_introduction}

\input{fs2_previous}

\input{fs3_method}

\input{fs4_dataset}

\input{fs5_results}
\input{fs6_conclusions}

\appendix
\section{Statistical toolbox}

\input{appendix_a}

\section{Examples}
\input{appendix_b}

\bibliography{fselect}

\end{document}

%% file: fs0_abstract.tex
\begin{abstract}
This paper describes a method for identification of the informative variables in the information system with discrete decision variables.  
It is targeted specifically towards discovery of the variables that are non-informative when considered alone, but are informative when the synergistic interactions between multiple variables are considered.  
To this end, the mutual entropy of all possible k-tuples of variables with decision variable is computed.  
Then, for each variable the maximal information gain due to interactions with other variables is obtained.  
For non-informative variables this quantity conforms to the well known statistical distributions.
This allows for discerning truly informative variables from non-informative ones. 
For demonstration of the approach, the method is applied to several synthetic datasets that involve complex multidimensional interactions between variables.  
It is capable of identifying most important informative variables, even in the case when the dimensionality of the analysis is smaller than the true dimensionality of the problem. 
What is more, the high sensitivity of the algorithm allows for detection of the influence of nuisance variables on the response variable. 
\end{abstract}

%% file: fs1_introduction.tex
\section{Introduction}

Modern datasets often contain very large number of variables, most of which are irrelevant for the phenomena under investigation. 
There are three main approaches to deal with such datasets. 
One can either select most relevant variables before modelling (filtering), use the dimensionality reduction techniques \citep{wall2003singular,roweis2000nonlinear,tenenbaum2000global,kim2003subsystem}
or apply modelling methodology that performs variable selection as a part of model building. 
The last option can be exercised in two flavours. 
Once can either use the modelling method with an 
embedded feature selection, or apply the wrapper method \citep{guyon2003introduction}. 
The well-known examples of the first approach are lasso \citep{lasso} or elastic network \citep{elastic_net} within domain of linear models.
The wrappers can be based on various machine learning algorithms, such as SVM \citep{cortes1995support} -- as in the SVM-RFE algorithm \citep{guyon2002gene}, or random forest \citep{Breiman2001} -- as in the Boruta algorithm \citep{Kursa2010a}.

The application of feature selection is not limited to building predictive models. 
One of its common applications is discovery of genes associated with diseases, either within the context of genome wide association studies (GWAS)\citep{hirschhorn2005genome}, or analysis of gene expression profiles \citep{alizadeh2000distinct}. 
The lists of genes revealed with statistical models are further analysed for their biological meaning and verified by wet-lab experiments. 
The number of variables analysed in gene expression studies can be as large as over fifty thousands \citep{haferlach2010clinical}, however the number is dwarfed by the GWAS where over ten millions of variables can be analysed \citep{fritsche2016large}. 
One should stress, that in both types of experiments the number of subjects is much lower than the number of variables. 
With the number of variables that high the computational cost of wrapper and embedded methods is so high that the initial filtering is de-facto standard approach, even if other methods are used later for building final models or construction of the sets of marker genes \citep{inza2004filter,mcshane2013criteria,haferlach2010clinical}. 

\subsection{Short review of filtering methods}

Numerous filtering approaches that have been proposed in the literature fall in two general classes. 
In the first one a simple measure of association between the decision variable and descriptive variables is computed, and the variables are ranked according this measure. 
The association can be measured using for example t-test for population means, signal to noise ratio, correlations between decision and descriptors, as well as various measures based on information entropy \citep{dudoit2002comparison,su2003rankgene}.  
The inclusion to the relevant set is decided by application of the statistical test using either FWER \citep{hochberg1988sharper}, or FDR \citep{benjamini_hochberg} method. 
In many cases the set is heuristically truncated to top $N$ features, when selection of $N$ is dictated by convenience \citep{jirapech2005feature}. 
The simple univariate filtering may overlook the variables that affect the investigated phenomenon only in interaction with other features.   
The number of overlooked but important features may be even higher when truncation is used, since some features that are not very relevant by themselves, may become very relevant when considered in conjunction with other features. 

The second class of filters uses both the measure of dependence between predictors and dependences within the set of predictor variables to obtain a small set of non-redundant variables that can be used to build predictive models. 
In some cases, the optimal subset is obtained in a two-step procedure, where in the first step the ranking is obtained and then the selection of non-redundant optimal feature set is performed \citep{zhao2007searching}. 
The interdependence between variables is examined using simple measures such as correlations between features \citep{hall2000correlation,yu2003feature}, measures based on the information theory \citep{zhao2007searching,navarro2009gene}, monotonous dependence  between variables \citep{bolon2011statistical} and various measures of dependency, relevance and redundancy \citep{kononenko1994estimating,peng2005feature,meyer2008information,wang2013selecting}.
Recently, the unifying approach based on the Hilbert-Schmidt Independence Criterion (HSIC) \citep{gretton2005measuring} has been proposed \citep{song2012feature}. It uses nonlinear kernels transforming the feature space and correlations to establish dependence between variables. 
It has been shown, that certain selection of kernels are equivalent to the well established methods. 

\subsection{Case for all-relevant feature selection}

The relevant variables, which are identified in the filtering, can be further used for building predictive models for phenomena under investigation. 
However, one should not confuse predictive ability with understanding the mechanisms. 
Statistical and machine learning models may achieve very high predictive power without understanding  the underlying phenomena. 
To obtain such a model, it is merely sufficient when descriptive variables and decision variables are influenced by some unknown hidden driver mechanism.  

On the other hand, the list of relevant variables can be useful for the researchers working in the field to formulate hypotheses, design experiments and uncover the mechanisms governing phenomena under scrutiny.  
We would like to argue that this list should be as complete as possible; at best it should contain all statistically relevant variables. 

There are at least three distinct reasons for that, which can be illustrated by the following example set in the context of gene selection. 
Let us assume the hypothetical situation when  simultaneous but modest increase of activity of three genes {\it a, b} and $c$ is required to represses activity of gene $D$, which is causing the phenotypic effect under scrutiny. 
In parallel the increased activity of these genes triggers the activity of the fourth gene $E$, whose expression is changed by orders of magnitude, but is unrelated to phenotypic effect.  
In this example the expression levels of genes $a$, $b$ and $c$ are the driver variables and expression levels of genes $D$ and $E$ are effector variables. 
The expression level of gene "D" is the decision variable, whereas the all remaining variables are descriptors. 

In this example, the mechanisms driving the effects under scrutiny are complex and weakly demonstrated, whereas the effects are strong and easy to see. 
The simple analysis of statistical strength of the effect will discover the strong association between $E$ and $D$, which may be sufficient to build the good predictive model. 
At the same time it may fail to discover any of the driver variables. 

Secondly, the measurements are inherently noisy and this influences the measures of relevance.  
An example of this effect is presented in the \ref{xmp_weaker}. 
For our token example, the subtle effects in driver variables $a$, $b$ and $c$ may be hidden in the noise, whereas the amplified effects in the effector variable $E$ can be clearly visible. 

Finally, the heuristic procedures that are used to build the optimal set may result in the selection of effector variables that are optimal for model building and drop the driver variables. 
In our token example the optimal set consist of variable $E$. 

The procedure that reveals relevance of all genes {\it a, b, c} along with easily discovered relevance of D would give the researcher whole information and could possibly lead to discovery of the driver mechanism. 

The univariate algorithms are simple and efficient and their result is an easy to understand list of variables which have a statistically significant connection to the decision variable. 
Unfortunately, they omit the variables that are relevant only in the interactions with other variables. 
Hence, in our token example, such algorithms would discover the variable $E$ only. 
Such variables can be identified by the heuristic multi-dimensional methods, however, in most cases only the subset of all relevant variables is reported - in our example the variable $E$. 

The example of the algorithm that both returns the full list of relevant variables and is able to discover variables that are involved in complex non-linear interactions is Relief-f \citep{kononenko1994estimating}. 
Unfortunately, it is biased against weaker and correlated variables \citep{robnik2003theoretical}. 
This problem may be specially important for example when multiple effector variables are strongly correlated with one driver variable, which is in turn correlated with the decision variable. 
The bias against highly correlated variables may result in removing an entire set of highly correlated variables from the final result. 
The bias against weaker predictor variables may lead to diminishing the estimate of their relevance to the point where they are non-separable from the irrelevant ones. 

\subsection{Current study}

The goal of the current study is to introduce a rigorous approach for the problem of the all-relevant variable selection, which also includes variables that are relevant due to non-trivial interactions with other variables, variables that are highly correlated and those that are weakly connected with the decision variable. 

To this end, we propose to use multidimensional exhaustive analysis of the mutual information between decision variable and descriptor variables. 
Such analysis requires both large number of experimental samples and massive computations that were not feasible until recently. 
However, rapid development of experimental technology as well as the growth of the computing power, especially using GPU for computation, 
allows for exhaustive search of pairs and triplets of variables  in the context of GWAS \citep{Goudey2013,hu2013information} as well as gene expression. 
Currently, even higher-dimensional interactions can be investigated in reasonable time for smaller number of variables.  

Nonetheless, the technological developments have not been matched by a methodological advances that are required to identify the synergistic interactions of multiple variables. 
In particular, it has been shown \citep{krippendorff,hu2013information} that the standard measure of synergy can give misleading results in higher-dimensional cases. 
The attempts to propose different measures proved unsatisfactory -- the tests reported significant multivariate interactions when there are none, or were unable to discover any interactions when they are doubtlessly present (see the next section for examples). 

The current study introduces a method of identifying variables for which synergistic interactions with other variables increase information about the response variable. 
It is based on the information theory and allows for precise estimates of the statistical importance of findings.
The method is tested using synthetic dataset with complex relations between descriptive variables, and with four various 3-dimensional response functions.

%% file: fs2_previous.tex
\section{Some previous approaches to the multivariate exhaustive search}

\subsection{Regression-based approach}

The regression-based tests are widely used for detection of the multivariate interactions
\citep{van2011travelling,vanderweele2010epistatic,kam2012glide}. An example of the regression model of interaction between 2 variables $X_1$, $X_2$, and the response variable $Y$ is a linear regression:
\begin{equation}
\label{regr}
Y=\alpha_0+\alpha_1 X_1 + \alpha_2 X_2 + \alpha_{12} X_1 X_2
\end{equation}
In the presence of the interaction, the empirically obtained value of $\alpha_{12}$ differs from 0 in a statistically significant amount. 

There are, however, some limits of the regression test for interactions. 
First, such a test is limited to detection of monotonous interactions only. 
This is not a serious issue in the case of continuous decision variables, since there are usually some monotonous components of interactions. 
The problem appears for discrete response variables and also possibly for continuous variable with strongly non-monotonous interactions. 
For discrete response variables, it has been shown by \citec{VanderWeele in}{vanderweele2010epistatic}, that the positive value of $\alpha_{12}$ is equivalent to a very strong probabilistic condition. 
Therefore, the regression-based test still can ignore some important interactions.
Moreover, the test becomes more complicated and less reliable, when more variables are considered. Then the regression model contains all the mixed terms -- e.g. the model, that describes interactions between 3 explanatory variables consists of 8 terms. The errors of coefficients grow rapidly, which reduces the sensitivity of the test in more than 2 dimensions. 

\subsection{Information measure of synergy}

Several approaches have been proposed for directly searching for synergies, using information theory measure \citep{boost,hu2013information,anastassiou2007}. 
The most known synergy measure, that uses information theory, is the interaction information of the variable set $\nu=\{Y,X_1,\ldots,X_k\}$:
\begin{equation}
I_{int}({\nu})=-\sum_{\tau\subseteq \nu}(-1)^{|\nu| - |\tau|} H(\tau) 
\end{equation}
where $\sum_{\tau\subseteq\nu}$ denotes summation over all subsets of $\nu$, and $H(\tau)$ is 
an entropy of the subset $\tau$ \citep{anastassiou2007}. The function becomes positive in presence 
of $k$-dimensional synergy, while less-dimensional interactions don't contribute the positive terms.

For a pair of variables $X_1$, $X_2$ the interaction information reads:
\begin{equation}
\label{syn2}
\begin{split}
I_{int}(Y,X_1,X_2)=&-H(Y,X_1,X_2)\\
  & +H(Y,X_1)+H(Y,X_2)+H(X_1,X_2)\\
  & -H(Y)-H(X_1)-H(X_2)\\
  \end{split}
\end{equation}
The interaction information is, however, difficult to use for reliable statistic tests.
The basic problem is connected with the fact, that the interaction information indicates not only synergy, but also  redundancy of the set of variables. 
The correlations between features add negative terms to $I_{int}(Y,\{X\})$. As a result,
\begin{itemize}
\item its distribution under the null hypothesis (i.e. in the absence of synergies) is not uniquely defined. 
It depends on the correlations between the explanatory variables;
\item in the extreme cases of statistically dependent variables, the interaction information can be zero or negative in spite of existing interactions. 
The unexpected behaviour of $I_{int}$ had been reported for subsets of 3 or more variables \citep{hu2013information,anastassiou2007}. 
The authors proposed some improvements that were claimed to fix the problem in these cases. 
However, the issue seems to be inherent, since the interaction information test may happen to fail even for 2 explanatory variables,  see Appendix~\ref{problem_synergy} for the illustrative example. 
\end{itemize}

\subsection{Multivariate test of association}

The problems with synergy measures led some researchers to examine an association between the response variable and the entire $k$-tuple of variables treated as a single feature. 
In such a case the features that are not relevant alone but show synergistic relations with other features can be identified as relevant. 
However, there is a significant risk of false positive results, since most $k$-tuples containing at least one feature, which is  correlated with the response variable, would be reported as relevant.  
In the case of high-dimensional datasets with numerous relevant variables, the number of reported $k$-tuples can be extremly large \citep{herold2009intersnp,Goudey2013}. 
One could possibly report only those $k$-tuples, where none of the variables has been recognised as relevant in the lower-dimensional search, but it is difficult to implement efficiently. 
What is more, such an approach would also overlook $k-$tuples with variables that are relevant in the univariate test, but are invloved in strong synergistic relations.

%% file: fs3_method.tex
\section{Searching for relevant variables}


The experiences described above inclined us to focus our attention on the variables and propose the method that identifies informative variables, including those that are not recognised as informative by a single-dimensional analysis. 
Once all the relevant variables are identified, the search of synergies will be much simpler. 
In particular, all the variables, which are identified as relevant in multi-dimensional analysis and not in the one-dimensional one, are relevant due to synergies. 

Identification of all informative variables can be conveniently described using the notion of {\em weak relevance}, introduced by \citec{Kohavi and John in}{kohavi_john}. 
\begin{definition}{weak relevance}\\
Let {\bf F} be a set of descriptors for the response variable Y. \\
Let {\bf $S_{X}$} be a set of all subsets {\it S} of {\bf F} that do not contain X.\\
The X is \emph{weakly relevant} if for any of S $\in$ {\bf $S_{X}$} the following relation is true: 
\begin{equation}
\label{weak_rel}
p(Y=y|X=x,S=s)\neq p(Y=y|S=s)
\end{equation}
\end{definition}
Hence, we can say that a variable $X$ is weakly relevant if there exists such a subset of variables $S$, that by adding variable $X$ to this subset increases information on the decision variable $Y$.   
In the same paper authors proposed also a higher level of relevance -- the strong relevance.  
A a variable $X$ is strongly relevant if for all possible subsets of variables $S$, adding variable $X$ to this subset increases information on the decision variable $Y$. 
\begin{definition}{strong relevance}\\
Let {\bf F} be a set of descriptors for the response variable Y. \\
Let {\bf $S_{X}$} be a set of all subsets {\it S} of {\bf F} that do not contain X.\\
The X is \emph{strongly relevant} if for all of S $\in$ {\bf $S_{X}$} the following relation is true: 
\begin{equation}
\label{strong_rel}
p(Y=y|X=x,S=s)\neq p(Y=y|S=s)
\end{equation}
\end{definition}

One may note, that these two notions are not sufficient to fully reflect all relationships between descriptive variables and decision variable. 
In particular, the weak relevance may reflect either correlation or synergy between descriptors.\\
To demonstrate the first case let's assume that two variables $X_1$ and $X_2$ are strongly correlated and inclusion of either of to the subset $S$ increases information on decision variable $Y$. 
However, when $X_1$ is already present then adding $X_2$ does not increase the information on $Y$ and vice-versa. 
The $X_1$ and $X_2$ are both weakly relevant.\\ 
For demonstration of synergy let's assume that $X_1$ and $X_2$ are both binary variables and that decision variable is also a binary variable. 
The decision variable  depends on both variables in a nonlinerar fashion: when $X_1=X_2$ the probability of $Y=1$ is increased by 10\%. 
Both variables are not correlated and  adopt value 0 or 1 with equal probability. 
When either variable is present in the dataset, then adding another increases the information on the decision variable. 
On the other hand, when both variables are absent, then adding either one does not increase the information. 
Again, by definition of weak relevance, both $X_1$ and $X_2$ are weakly relevant.

In some cases these definitions lead to counterintuive results. 
Consider the information system consisting of three descriptive variables $X_1, X_2, X_3$ and decision variable $Y$, where $X_1 = Y$, $X_2$ is a pure random noise, and $X_3 = X_1 + X_2$. 
It is easy to see, that a subset $S$ consisting of $\{X_2,X_3\}$ contains all information on $Y$, hence adding $X_1$ cannot add any information. 
On the other hand, adding any variable to $X_1$ also cannot add information on $Y$, hence all three variables are weakly relevant, despite that $X_3$ is clearly redundant and $X_2$ by itself has no information on decision variable. 

Yu and Liu \cite{yu2004efficient} proposed to split weak-relevance into two classes. 
In their approach variables can be {\it weakly relevant non-redundant} or  {\it weakly relevant redundant}. 
The redundant variables are defined using the notion of a Markov blanket \citep{koller1996toward}. 

\begin{definition}{Markov blanket}\\
Let {\bf $F$} be a set of descriptors for the response variable $Y$. \\
Given the variable $X$, let $M_X \in F$, $X \notin M_X$. $M_X$ is a Markov blanket for $X$ if 
\begin{equation}
\label{Markov_blanket}
p(F - X - M, Y |X,M_X) = p(F - X - M, Y |M_X)
\end{equation}
\end{definition}

\begin{definition}{Redundant variable}\\
A variable $X \in G$ is redundant in $G$ when there exists a Markov blanket for $X$ in $G$. 
\end{definition}

With this definition the minimal-optimal feature selection problem becomes a problem of finding all non-redundant variables. 
The notion of {\it weakly-relevant and redundant} variables helps to deal with the last example, since the $X_2$ and $X_3$ variables are clearly redundant. 
Unfortunately, it is not sufficient to help with other limitations. 

The definitions of weak and strong relevance are absolute in the sense that they require examination of the all possible subsets $S$  of the feature set $F$, disjoint with $X$, either to exclude the weak relevance or to prove the strong relevance of the variable $X$.  
An exhaustive search of all possible combinations is in most cases not possible due to limited computational resources and limited dataset. 

Nevertheless, despite all these shortcomings, the distinction between weak and strong relevance is very useful concept, that explicitly demonstrates that relevance is not a unique property, but can be graded. 
In the context of practical applications weaker and more specific definitions of weak and strong relevance may be more appropriate than very general definitons proposed by Kohavi and John.  
In particular, we propose to introduce two notions, {\it $k$-weak relevance} and {\it $k$-strong relevance} that better reflect our limited ability to test exhaustively possible combinations of variables. 

The variable $X$ is {\it $k$-weakly relevant} if its relevance can be established by analysing all $\binom nk$ subsets of $k$ variables that include variable $X$. 
More formally:  
\begin{definition}{$k$-Weak Relevance}\\
Let {\bf F} be a set of descriptors for the response variable Y. \\
Let {\bf $S_{k-1;X}$} be a set of all subsets {\it S} of {\bf F} with cardinality $(k-1)$ that do not contain X.\\
The X is \emph{$k$-weakly relevant} if for any of S $\in$ {\bf $S_{k-1;X}$} the following relation is true: 
\begin{equation}
\label{k-weak_rel}
p(Y=y|X=x,S=s)\neq p(Y=y|S=s)
\end{equation}
\end{definition}

Similarly, the variable $X$ is {\it $k$-strongly relevant} if its it strongly relevant in all $\binom nk$ subsets of $k$ variables that include variable $X$. 
More formally:  
\begin{definition}{$k$-Strong Relevance}\\
Let {\bf F} be a set of descriptors for the response variable Y. \\
Let {\bf $S_{k-1;X}$} be a set of all subsets {\it S} of {\bf F} with cardinality $(k-1)$ that do not contain X.\\
The X is \emph{$k$-strongly relevant} if for all of S $\in$ {\bf $S_{k-1;X}$} the following relation is true: 
\begin{equation}
\label{k-strong_rel}
p(Y=y|X=x,S=s)\neq p(Y=y|S=s)
\end{equation}
\end{definition}

One may note, that the standard univariate filtering is equivalent to finding all $1$-weakly relevant variables.
In this case the subset $S$ in the definition~\ref{k-strong_rel} is simply the empty set $\emptyset$.  

Selection of the particular value of $k$ depends on the available computational resources and size of the dataset. 
One should note, that even with infinite computational resources the analysis in very high dimensions can be applied only for sufficiently large samples, since 
good representation of the probability density is necessary to obtain reliable results. 
Therefore in practice the dimensionality of the analysis may be limited both by the computational resources, when the number of variables is high, or by the sample size, when the number of objects is small. 
The presented method of feature selection is designed as an initial step of investigation. 
The analysis of the redundancy and synergies between variables is not performed here and left for the next step of the analysis.  

To obtain a list of all variables, that exhibit the statistically significant influence on a response variable we use the conditional mutual information between $Y$ and $X$ given the subset $S$:
\begin{equation}
\label{condig}
  I(Y;X|S)=\left[H(X,S)-H(Y,X,S)\right]-\left[H(S)-H(Y,S)\right]
\end{equation}
where $H(X)$ denotes the information entropy of the variable. 

$I(Y;X|S)$ can be interpreted directly as amount of information about the response variable $Y$, that is contributed by the variable $X$ to the subset $S$. 
It is always non-negative and becomes zero, when the variable contributes no information to the subset.

Note that in the univariate case, i.e. if the subset $S$ is empty, $I(Y;X|S)$ reduces to the mutual information of $Y$ and $X$, commonly used to test the statistical association between the variables.
\[I(Y;X|\emptyset)=I(Y;X)\]
The conditional mutual information has been used in the context of minimal-optimal feature selection, see for example \citep{peng2005feature,herold2009intersnp,navarro2009gene,vergara2014review}, however it has not been used for identification of the synergistic relevant variables.

\subsection{The statistics of conditional mutual information}

Computation of conditional 
mutual information is straightforward and efficient for discrete variables. 
In the case of continuous variables there are two possibilities. 
One can either build an estimate of a multidimensional probablity density function, or one can discretize variables.   
In the current study we use the quantile discretisation, which method is both simple and robust. 
To avoid overfitting, the discretisation is not well aligned with the test functions - the descriptive variables are discretised into three equipotent categories, whereas the decision variable is based on power of two periodicity. 
 


Let the response variable $Y$, the tested variable $X$ and the variables $\{S_i\}$,~$i=1,\ldots,k-1$, be the discrete ones with number of categories $C_Y$, $C_X$, $\{C_{S_i}\}$, respectively. 
The analysed dataset contains values of the variables for $N$ objects. 
The obvious estimate of the probability, that the variable takes on a particular value, is
\[\hat{p}_x=\frac{n_x}{N}\]
where $n_x$ is the number of objects in the dataset, for which the variable $X$ takes on the value $x$. 
Hence, the estimate of information entropy reads:
\[\hat{H}(X)=-\sum_x \hat{p}_x \log\hat{p}_x \]
and the estimate of the conditional mutual information is equal to:
\begin{equation}
 \label{estcond}
 \begin{split}
  \hat{I}(Y;X|S)&=\left[\sum_{y,x,\{s_i\}}\hat{p}_{yx\{s_i\}}\log \hat{p}_{yx\{s_i\}}
    -\sum_{x,\{s_i\}}\hat{p}_{x\{s_i\}}\log \hat{p}_{x\{s_i\}}\right]\\
    &-\left[\sum_{y,\{s_i\}}\hat{p}_{y\{s_i\}}\log \hat{p}_{y\{s_i\}}
    -\sum_{\{s_i\}}\hat{p}_{\{s_i\}}\log \hat{p}_{\{s_i\}}\right]\\
 \end{split}
\end{equation}
where
\begin{description}
\item{ $\hat{p}_{yx\{s_i\}}=\frac{n_{yx\{s_i\}}}{N}$, } $n_{yx\{s_i\}}$ is the number of objects for which \[Y=y,\;X=x,\;S_i=s_i,\]
\item{ $\hat{p}_{x\{s_i\}}=\frac{n_{x\{s_i\}}}{N}$,} $n_{x\{s_i\}}=\sum_y n_{yx\{s_i\}}$, 
\item{ $\hat{p}_{y\{s_i\}}=\frac{n_{y\{s_i\}}}{N}$,} $n_{y\{s_i\}}=\sum_x n_{yx\{s_i\}}$, 
\item{ $\hat{p}_{\{s_i\}}=\frac{n_{\{s_i\}}}{N}$,} $n_{\{s_i\}}=\sum_{y,x} n_{yx\{s_i\}}$. 
\end{description}

If $X$ contributes no information about the response variable to the subset $S\equiv\{S_i\}$ and the sample size $N$ is sufficiently large, then $2 N \hat{I}(Y;X|S)$ follows 
the well-known $\chi^2$ distribution, see Appendix~\ref{df_cond} for details:
 \begin{gather}
 2 N \hat{I}(Y;X|S) \sim \chi^2(df),\\
 df=(C_Y-1)(C_X-1)\prod_{i=1}^{k-1} C_{S_i} 
 \end{gather}

Some examples of behaviour of conditional mutual information for various cases of variable associations are shown in Fig. \ref{xmp_rel}, \ref{xmp_synr} in the Appendix~\ref{xmp_weaker}. 

\subsection{The minimum $p$-value analysis}

For each value of $\hat{I}(Y;X|S)$, one can compute the associated $p$-value and determine the probability, that the variable $X$ contributes no information to the subset $S$.
To check the $k$-weak relevance of the variable, we should determine whether there exists any subset $S$ of $k-1$ features, in context of which $X$ is relevant. 
To this end we find the minimum $p$-value over all the subsets for each variable: 
\begin{equation}
\label{pmin}
p_{min}(X)=\min_{S\subset F}\left[ p_{\chi^2}\left(\hat{I}(Y;X|S)\right)\right].
\end{equation}
If the variable $X$ is irrelevant, $p_{min}(X)$ proves to follow the exponential distribution, for details, see Appendix~\ref{statmax}:

\begin{equation}
\label{expmin}
P\left(p_{min}(X)<v\right)=1-e^{-\gamma v}.
\end{equation}

The parameter $\gamma$ can be estimated based on the test results for all the variables, most of which are irrelevant. This allows us to calculate the probability of being irrelevant (the $p$-value) for each variable $X$. The result can be used directly to decide, whether the variable should be selected as relevant, via standard FWER (e.g. Bonferroni correction) or FDC (like \citec{Benjamini-Hochberg}{benjamini_hochberg}) methods.

\section{The algorithm}

The theory described above leads to the algorithm of multivariate feature selection. 
It consists of the following steps:
\begin{itemize}
\item If the response variable and the explanatory variables are continuous, they are discretised first.
\item For each $k$-tuple of variables $S^k$ the contingency table is built, that contains number of objects, for which the variables take on particular values. 
This step is the most expensive computationally, however, it is simple, so it can be performed in parallel, for example using GPU. 
\item For each variable $X\in S^k$ the conditional mutual information between the response variable $Y$ and $X$ given the rest of the $k$-tuple $S^{k-1}$ is calculated. Then, the associated $\chi^2$~$p$-value is calculated.
\item For each variable the minimum $p$-value over all the $k$-tuples is recorded as $p_{min}(X)$.
\begin{quote}
In a special case, when all the explanatory variables have the same number of categories (e.g. they are all the discretised continuous variables), the last two steps can be simplified. 
In this case, the minimum $p$-value corresponds to the maximum value of $I(Y;X|S^{k-1})$. 
Instead of computing all the $p$-values, it is enough to record the maximum $I(Y;X|S^{k-1})$, then calculate the associated $p$-value once for a variable.
\end{quote}
\item Based on the statistics of $p_{min}$ over all the variables, the parameter $\gamma$ in Eq.~(\ref{expmin}) is estimated (see Appendix~\ref{statmax}).
\item The eventual $p$-value is calculated for each variable.
\item Finally, either Bonferroni-Holm FWER method, or Benjamini-Hochberg FDC method is used to decide which variables should be selected as relevant. 
\end{itemize}

%% file: fs4_dataset.tex
\section{Tests on Synthetic Datasets}

\begin{figure}[t]
\centering
 \begin{tabular}{ll}
 {\bf a)}& {\bf b)}\\
\includegraphics[width=0.48\textwidth]{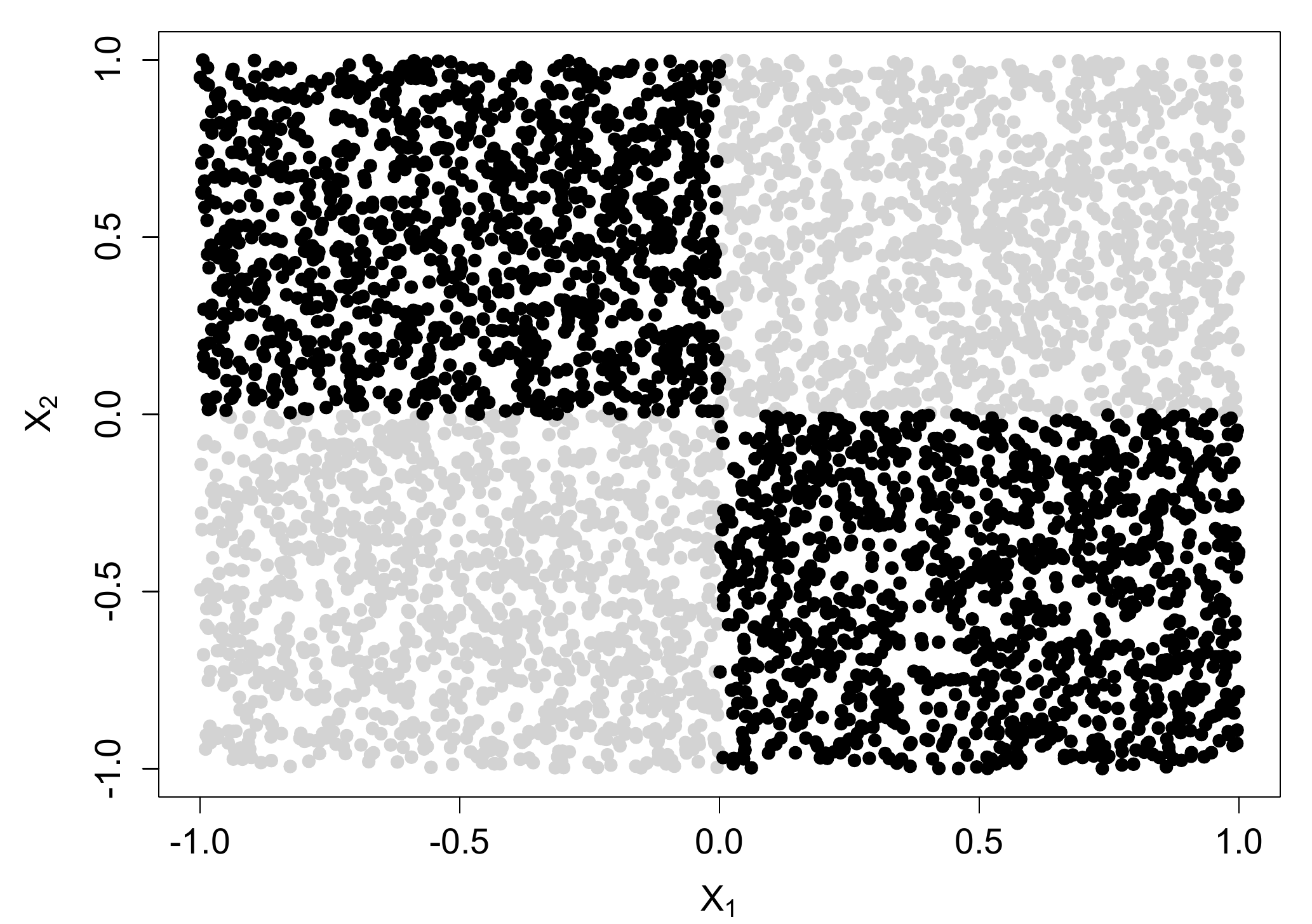}&
\includegraphics[width=0.48\textwidth]{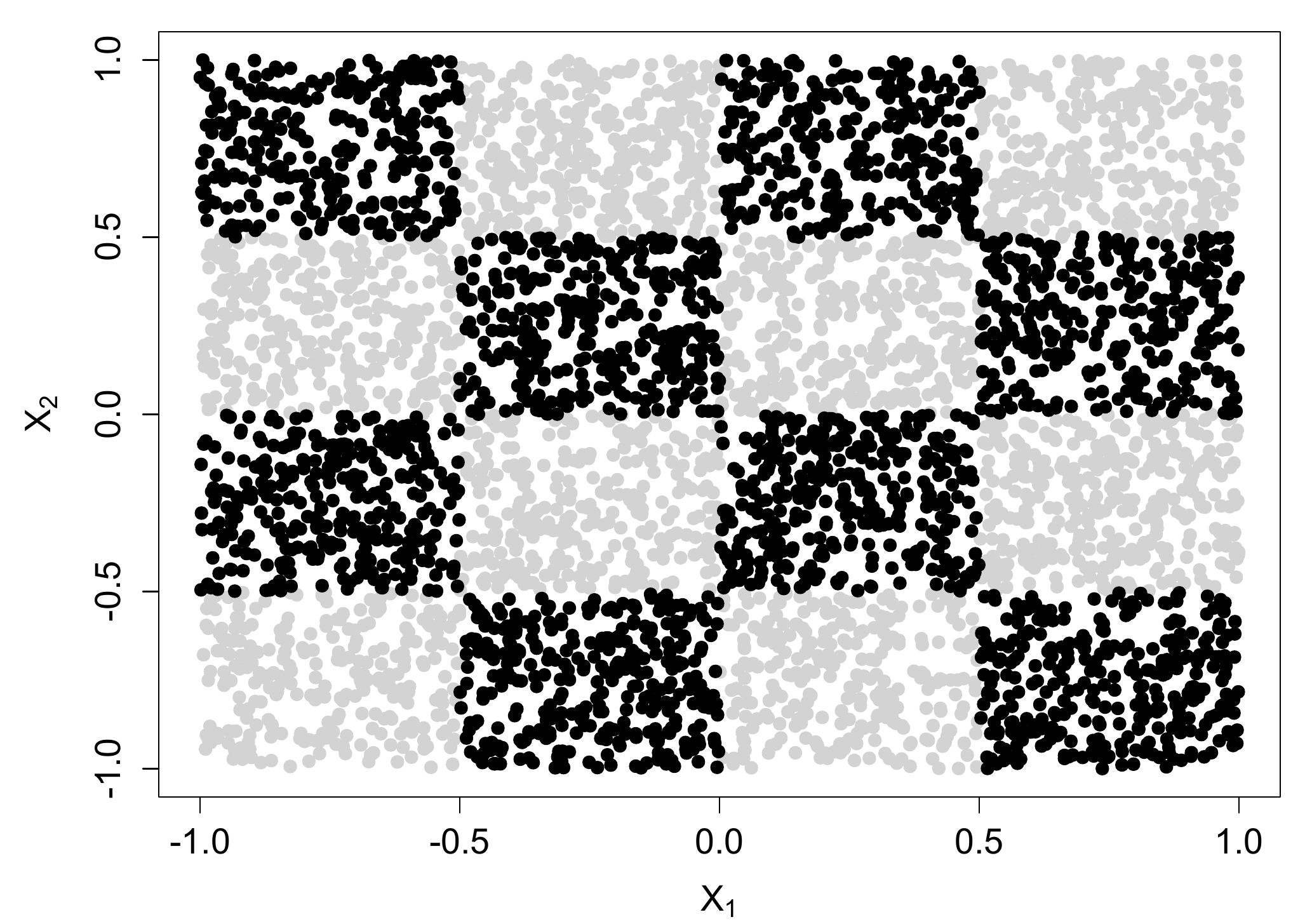}\\
 {\bf c)}& {\bf d)}\\
\includegraphics[width=0.48\textwidth]{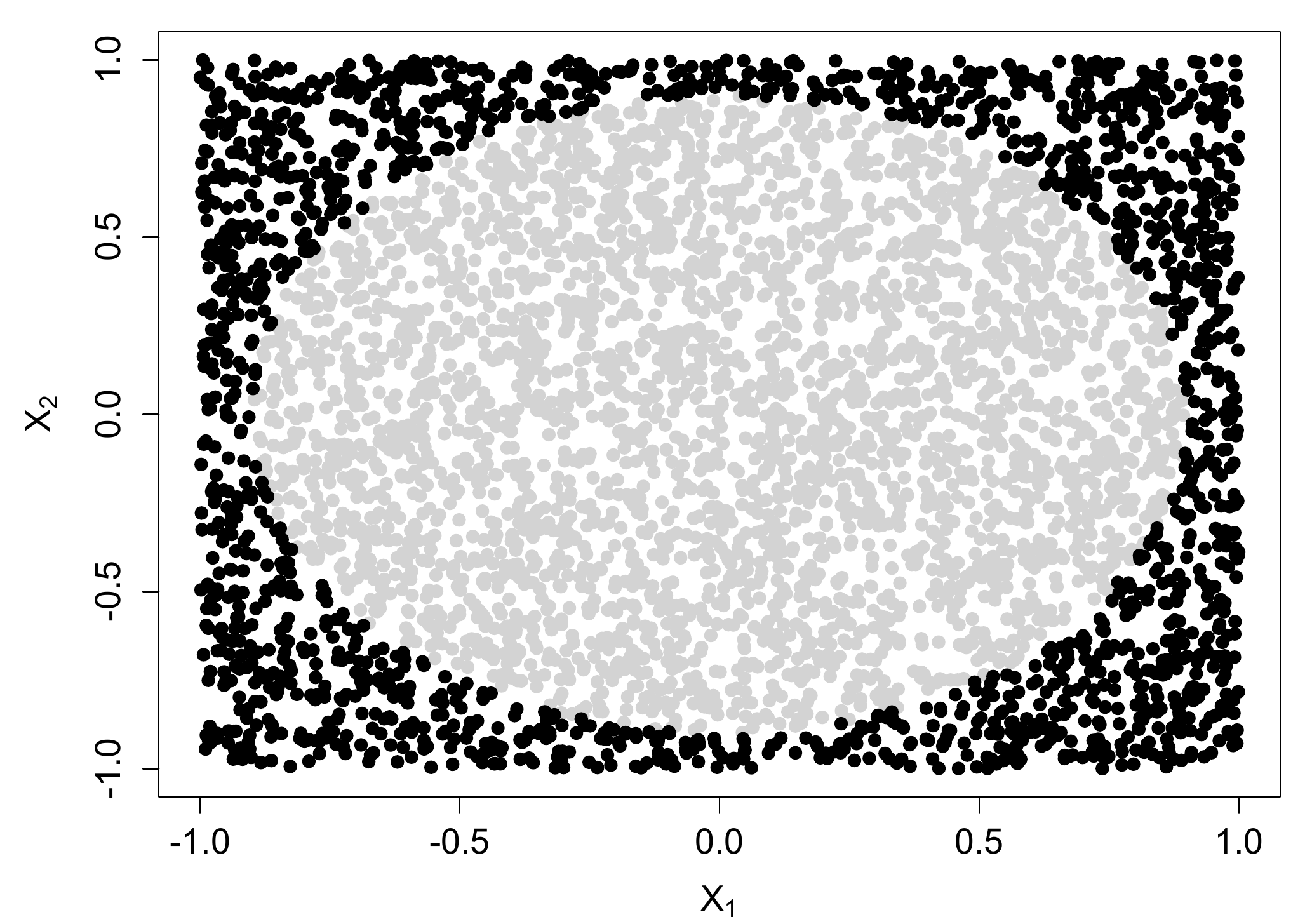}&
\includegraphics[width=0.48\textwidth]{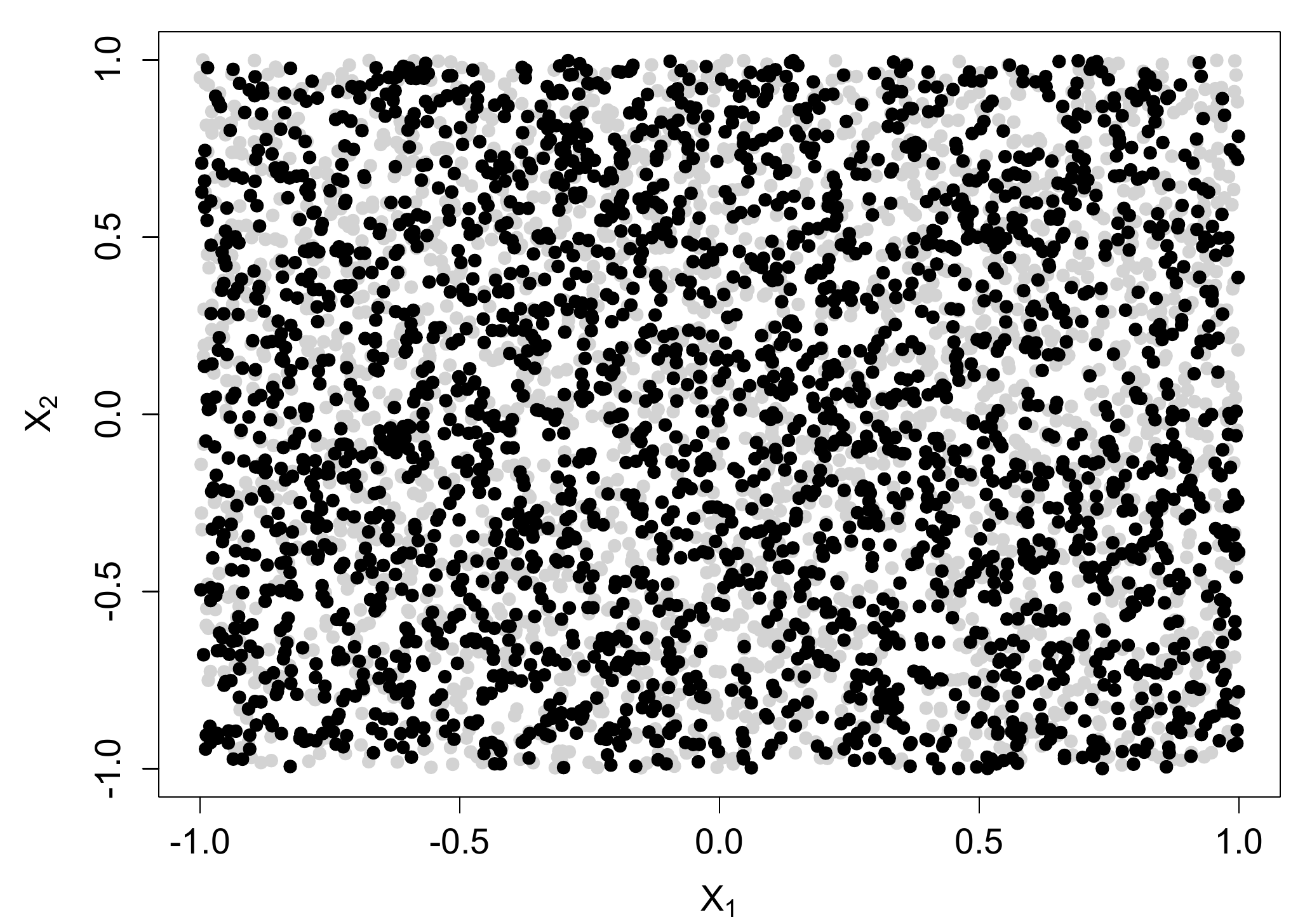}\\
\end{tabular}
\caption{2-dimensional illustration of the test dataset for various response variables:
 a) XOR: $Y=( X_1\cdot X_2<0 )$,
 b) checkerboard: $Y=( \sin(2 \pi X_1) \cdot \sin(2 \pi X_1)<0 )$, 
 c) sphere: $Y=( X_1^2+X_2^2>0.9 )$,
 d) random response.
 }
\label{y_2d} 
\end{figure}

To explore the performance of the method, we have generated a complex artificial dataset, containing 351 descriptive variables, defined for 5000 objects. 
There are two general classes of variables in this dataset. 
The first one consists of variables which are drawn from a $(-1,1)$ uniform distribution, with possible admixture of random noise.  
The second is obtained as linear combination of primary variables, also with possible admixture of random noise. 
The random noise is drawn from the $(-0.15,0.15)$ uniform distribution. 

More specifically, the dataset consists of the following groups of variables:
\begin{enumerate}
\item 3 base variables drawn from the uniform distribution. The response variable is a function of these variables only.
\item 3 base variables with 15\% random noise added. 
\item 20 linear combinations of the base variables (belonging to group 1) with random coefficients.
\item 20 linear combinations of the base variables (belonging to group 1) and nuisance variables (which are henceforth refferred to as group 5) with random coefficients and 15\% random noise added.
\item 5 nuisance variables drawn from the uniform distribution, see Appendix~\ref{xmp_multi}.
\item 100 random variables.
\item 200 linear combinations of 10 variables from group 6 with 15\% additional random noise.
\end{enumerate}
The variables from the groups 1-5 are expected to be weakly relevant, while groups 6-7 contain irrelevant ones. 
Obviously, the most important variables belong to the groups 1 and 2. 
The set simulates real datasets, where the pure signal is usually not available, and there are many irrelevant
variables that are correlated with each other. 

The response variable $Y$ is a binary function of first 3 variables. 
We tested 4 example functions:
\begin{itemize}
\item $Y=( X_1\cdot X_2 \cdot X_3 <0 )$
\item $Y=( \sin(2 \pi X_1) \cdot \sin(2 \pi X_2) \cdot \sin(2 \pi X_3) <0 )$
\item $Y=( X_1^2+X_2^2+X_3^2>0.9 )$
\item $Y$ random, independent of $X_1$, $X_2$, $X_3$
\end{itemize}
2-dimensional versions of the functions are shown in Fig. \ref{y_2d}.

The methodology presented in the current study is general and applies both to continuous and discrete variables, but for simplicity and efficiency it is restricted here to  the discrete case. 
Therefore, the descriptive variables were discretised to conduct the tests. 
In all cases the variables were discretised into 3 equipotent monotonic categories. 
The discretisation was chosen with three criteria in mind. 
Firstly, it does not give unfair advantage to the base functions, since the discretisation is not particularly well fitted to the response functions under consideration. 
Secondly, it assures that the representation of each category in multidimensional cases is sufficiently large to minimise variance due to sampling. 
Finally, the single predefined split gives insight in the influence of the split on the results, at least in the case of the base variables. 
Statistical tests for discovery of explanatory variables were performed for each response variable, using FDR procedure with $\alpha=0.1$ with one-, two- and three-dimensonal analyses. 

The results of a feature selection procedure 
are often used to build predictive models of the phenomena under scrutiny, with the underlying assumption, that the most relevant (most informative) variables should be used for model building. 
In the case of the current study we have tested, whether variables selected by multidimensional feature selection procedure lead to better models than those obtained using univariate analysis. 
To this end, the variables selected by the procedure were used to build a predictive models using Random Forest classifier \citep{Breiman2001} implemented in R package randomForest \citep{RManual,Liaw2002Classification}. 
The procedure was performed either with three highest scoring variables or with all variables deemed relevant by the algorithm.  
As a reference we used predictive models built using following subsets of relevant variables: 
\begin{itemize}
\item{all variables,}
\item{all relevant variables,}
\item{pure base variables}
\item{pure combination variables}
\item{pure base and combination variables}
\end{itemize}

%% file: fs5_results.tex
\section{Results}


The three response functions that were generated for the same set of descriptive variables represent a wide range of difficulties - the easy problem in the case of sphere, intermediate in the case of exclusive or, and the 3D checkerboard is the most difficult, as can be seen in the Table~\ref{found_table1}, where the number of variables identified as relevant is reported. 

\begin{table}
\centering
\caption{The number of variables identified as relevant for all four response variables obtained in 1-, 2- and 3-dimensional analysis. The FDR procedure with $\alpha=0.1$ was used in all cases to call relevant variables. Seven categories of variables as labelled as follows: G1 -- base variables, G2 -- base variables distorted with 15\% random noise, G3 -- random linear combinations of base variables, G4 -- random linear combinations of base variables distorted with 15\% of random noise and nuisance variables with the same amplitude, G5 -- nuisance variables, G6 -- uncorrelated random variables, G7 -- linear combinations of G6. }
{\addtolength{\tabcolsep}{0pt}
\begin{tabular}{c|cc|cc|c|cc}
variable group	&	G1	&	G2	&	G3 	&	G4 	&	G5 	&	G6	&	G7	\\
\# of variables	&	3	&	3	&	20	&	20	&	5	&	100	&	200	\\
\hline
\multicolumn{8}{c}{3D sphere} \\	
1d		&	3	&	3	&	20	&	8	&	0	&	0	&	0	\\
2d		&	3	&	3	&	20	&	20	&	4	&	0	&	4	\\
3d		&	3	&	3	&	20	&	20	&	5	&	0	&	3	\\
\hline
\multicolumn{8}{c}{3D XOR} \\
1d		&	0	&	0	&	10	&	2	&	0	&	1	&	0	\\
2d		&	3	&	3	&	20	&	20	&	3	&	1	&	1	\\
3d		&	3	&	3	&	20	&	20	&	5	&	1	&	2	\\
\hline
\multicolumn{8}{c}{3D checkerboard (product of sines)} \\	
1d		&	0	&	0	&	2	&	0	&	0	&	0	&	0	\\
2d		&	2	&	2	&	20	&	2	&	0	&	2	&	6	\\
3d		&	3	&	3	&	20	&	4	&	0	&	1	&	3	\\
\hline
\multicolumn{8}{c}{Random response} \\	
1d		&	0	&	0	&	0	&	0	&	0	&	0	&	0	\\
2d		&	0	&	0	&	0	&	0	&	0	&	0	&	0	\\
3d		&	0	&	0	&	0	&	0	&	0	&	0	&	0	\\
\hline
\end{tabular}
}
\label{found_table1}
\end{table}

It can be seen, that while the original problems are formulated in three dimensions, the presence of linear combinations of variables can reduce the apparent dimensionality of the problem. 

\subsection{An easy problem - a 3D sphere}
\begin{figure}
\centering
\begin{tabular}{ccc}
1D  & 2D & 3D \\
\includegraphics[width=0.28\textwidth]{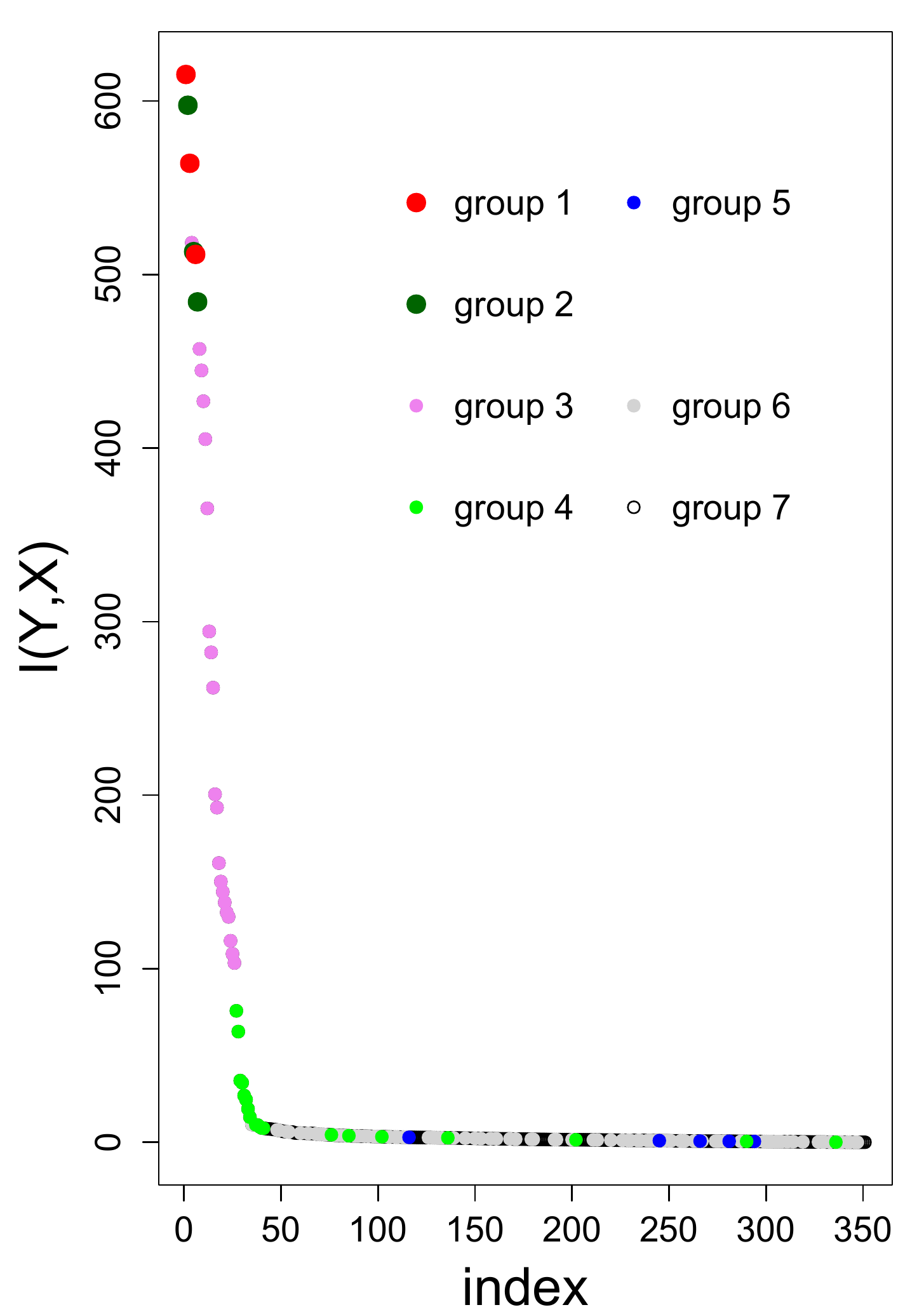}&
\includegraphics[width=0.28\textwidth]{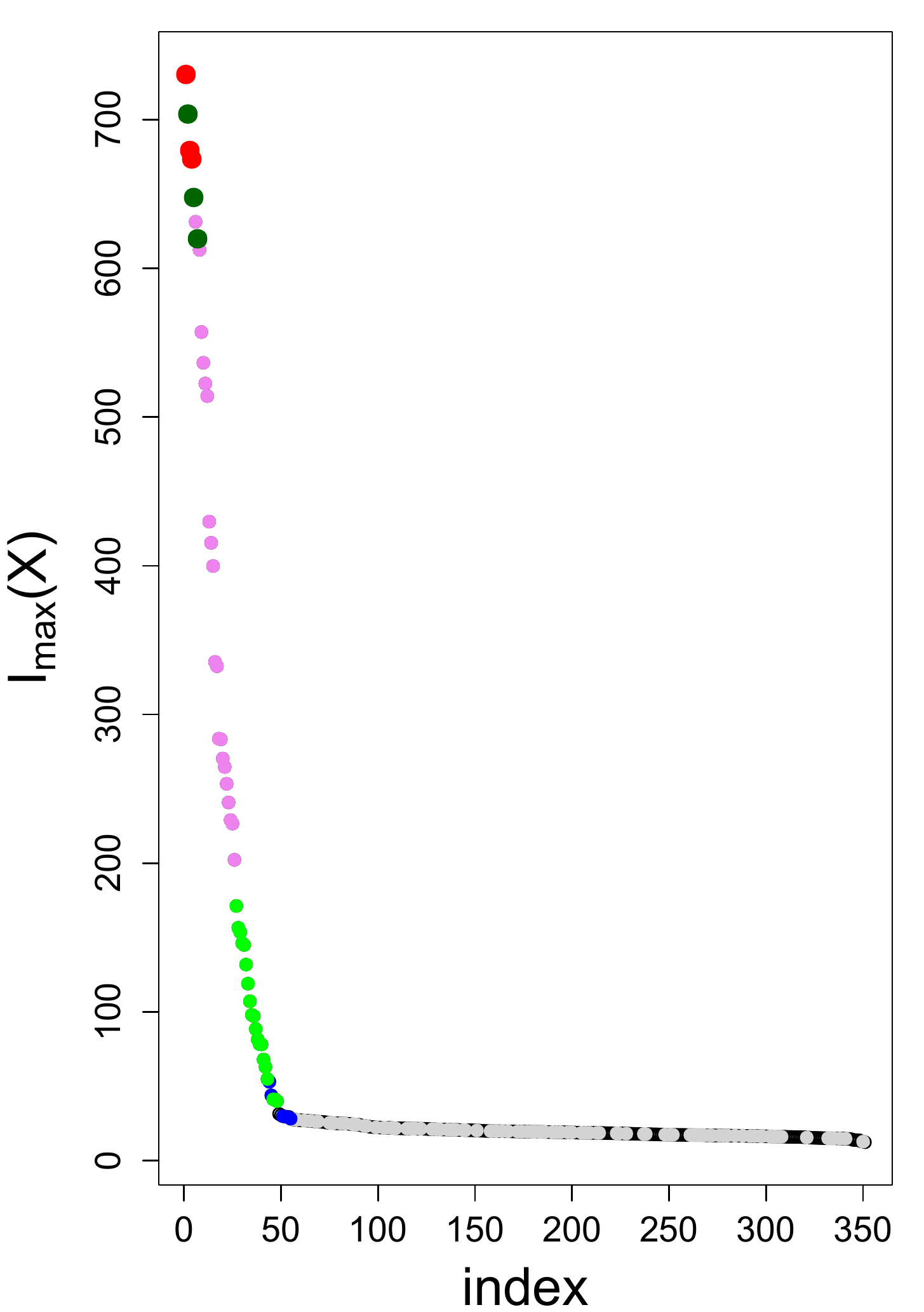}&
\includegraphics[width=0.28\textwidth]{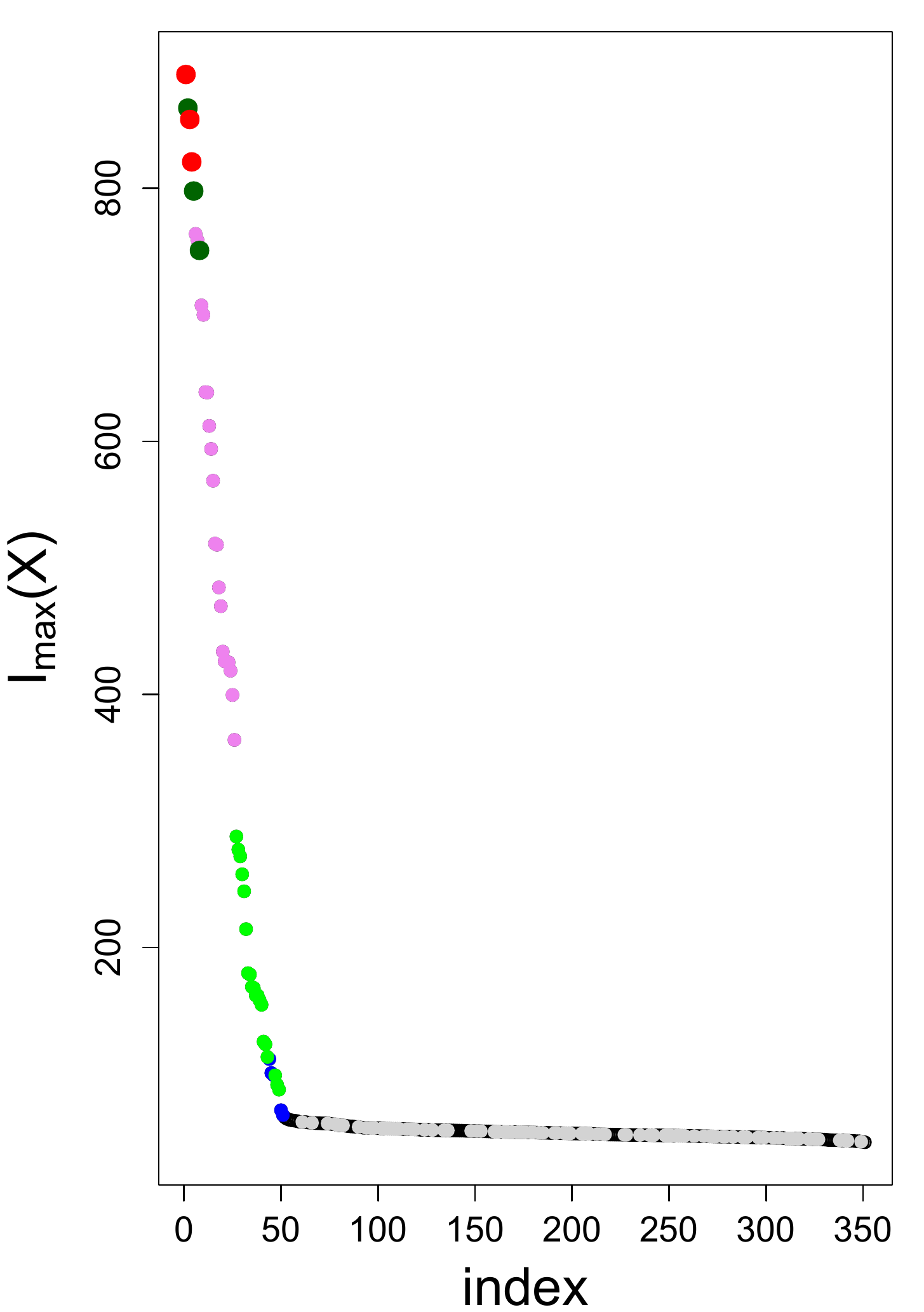}\\
\end{tabular}
\caption{Information gain for 1-, 2- and 3-dimensional analysis of relevance for response variable defined as $Y=( X_1^2+X_2^2+X_3^2>0.9 )$ (3-dimensional sphere). 
}
\label{SphereRanks}
\end{figure}

In particular, in the case of the spherical decision function, most of the informative variables are easily identified even in one dimension. 
Extension of the analysis to higher dimensions does not lead to large changes of the information gain, see Figure~\ref{SphereRanks}. Additionally, the relative ranking of the groups of variables is mostly preserved. 
The ranking of variables is concordant with the intuition - the pure base variables are scored higher than noisy base variables that are in turn scored higher than linear combinations of pure variables. 
The noisy linear combinations are below pure linear combinations, and the nuisance variables are scored lowest among informative variables. 
Nevertheless, increasing the dimensionality of analysis allows for higher sensitivity in the case of noisy linear combination and nuisance variables.
In particular, transition from the 1- to 2-dimensional analysis improves the ranking of noisy linear combinations and nuisance variables, see  Table~\ref{ranks_table1} and Fig.~\ref{SphereRanks}. 
One should note that nuisance variables are informative only due to interactions with the noisy linear combinations, hence they are non-informative in one dimension and the increase of their ranking in higher dimensional analysis is expected. 

\begin{table}
\centering
\caption{The summary for ranks for variables is shown for 7 categories of variables. Categories are identical as in Table~1. For   pure and noisy base variables all ranks are shown, for remaining informative variables the highest and lowest rank, whereas for the random variables only the rank of the highest scoring variable is shown.}
{
\addtolength{\tabcolsep}{-2pt}
\begin{tabular}{c |  c c | c c |c| c c }
variable group	&	G1	&	G2	&	G3 	&	G4 	&	G5 	&	G6	&	G7	\\
\# of variables	&	3	&	3	&	20	&	20	&	5	&	100	&	200	\\
\hline
\multicolumn{8}{c}{3D sphere} \\	
1d & 1 3 6     & 2 5 7      & 4 26   & 27 336 & 116 294 & 35 & 42 \\
2d & 1 3 4     & 2 5 7      & 6 26   & 27  48 &  44  55 & 56 & 49 \\
3d & 1 3 4     & 2 5 8      & 6 26   & 27  49 &  44  51 & 61 & 52 \\
\hline
\multicolumn{8}{c}{3D XOR} \\
1d & 26 46 251  & 25 102 297 & 1 283 & 10 306 & 71 126 333 & 16 & 13 \\
2d & 7 18 26    & 8  15 25   & 1  24 & 27  48 & 45  49 301 & 50 & 51 \\
3d & 6 16 19    & 5  15 20   & 1  26 & 27  46 & 47  49  51 & 54 & 52 \\
\hline
\multicolumn{8}{c}{3D checkerboard (product of sines)} \\	
1d & 82 103 318 & 90 148 271 & 1 326 &  6 350 & 17  302  &  3 &  4 \\
2d & 23 28 57   & 19 29 94   & 1  24 & 14 328 & 88  323  & 33 & 25 \\
3d & 20 24 26   & 22 23 25   & 1  21 & 27 312 & 129 294  & 32 & 30 \\
\hline
\end{tabular}
}
\label{ranks_table1}
\end{table}

\subsection{Intermediate difficulty - a 3D XOR}
 
Three dimensional XOR is an example of a problem for which the multi-dimensional analysis and all-relevant feature selection is required for full understanding of the system under scrutiny.  
When only 3 base variables are taken into account, it is an example of pure 3-dimensional synergy, where no variable is informative unless all 3 are analysed together.  
However, the introduction of linear combinations of the base variables to the descriptive variables significantly affects the analysis. 

The one-dimensional analysis cannot discover importance of the base variables.
On the other hand, half of the pure linear combinations of base variables along with 2 noisy ones are deemed relevant, see Table~\ref{found_table1}.
Hence, the univariate approach to this problem reveals set of twelve variables that can be further used for building predictive models or to design marker sets. 
However, this set does not contain any of the original variables that were used to generate the problem, neither pure nor noisy ones. 
Therefore, any reasoning, directed towards elucidation of the mechanism for generation of the response variable, based on the relevant variables identified by the univariate analysis would be seriously flawed. 

Increasing the dimensionality of analysis to 2D is sufficient to identify nearly all relevant variables, only 2 out of 5 nuisance variables were not discovered.    
Nevertheless, the ranking of the variables does not reflect the true relevance of the variables. 
In particular, five highest scoring variables belong to the pure linear combination class, and the least informative pure base variable has the rank 25. 
Therefore, any analysis based on the highest scoring variables only, would also likely miss some of the truly relevant variables. 
Finally, the 3-dimensional analysis reveals all relevant variables, with slightly improved ranks of the base variables. 

It is interesting to see how the relative ranking of different classes of variables is affected by the dimensionality of the analysis, see Table~\ref{ranks_table1}.
In the one-dimensional analysis the highest scoring variables are linear combinations of base variables, with non-noisy versions consistently scored higher.  
The ranks of base variables are essentially randomly dispersed between 25 and 297.  
In two dimensions the relative ranking between groups of variables is radically changed. 
The top ranks still belong to non-noisy linear combinations of base variables.
However, the base variables themselves, both pure and distorted, appear much higher in the ranking, well within the range spanned by the linear combinations of pure base variables. 
Interestingly, the noisy combinations score significantly worse, with large gap between information gain of the pure and noisy combinations. 
Nonetheless, all noisy combinations were also identified as relevant, along with 3 of 5 nuisance variables. 
The ranking of base variables is even slightly better in the 3-dimensional analysis.
Yet, the highest rank for the base variable is only 5, and the lowest one is as far as 20.  
\begin{figure}
\centering
\begin{tabular}{cccc}
 & $I_{max}(X)$ & $p$-value & FDR \\
\raisebox{0.2\textwidth}{1D}&
\includegraphics[width=0.28\textwidth]{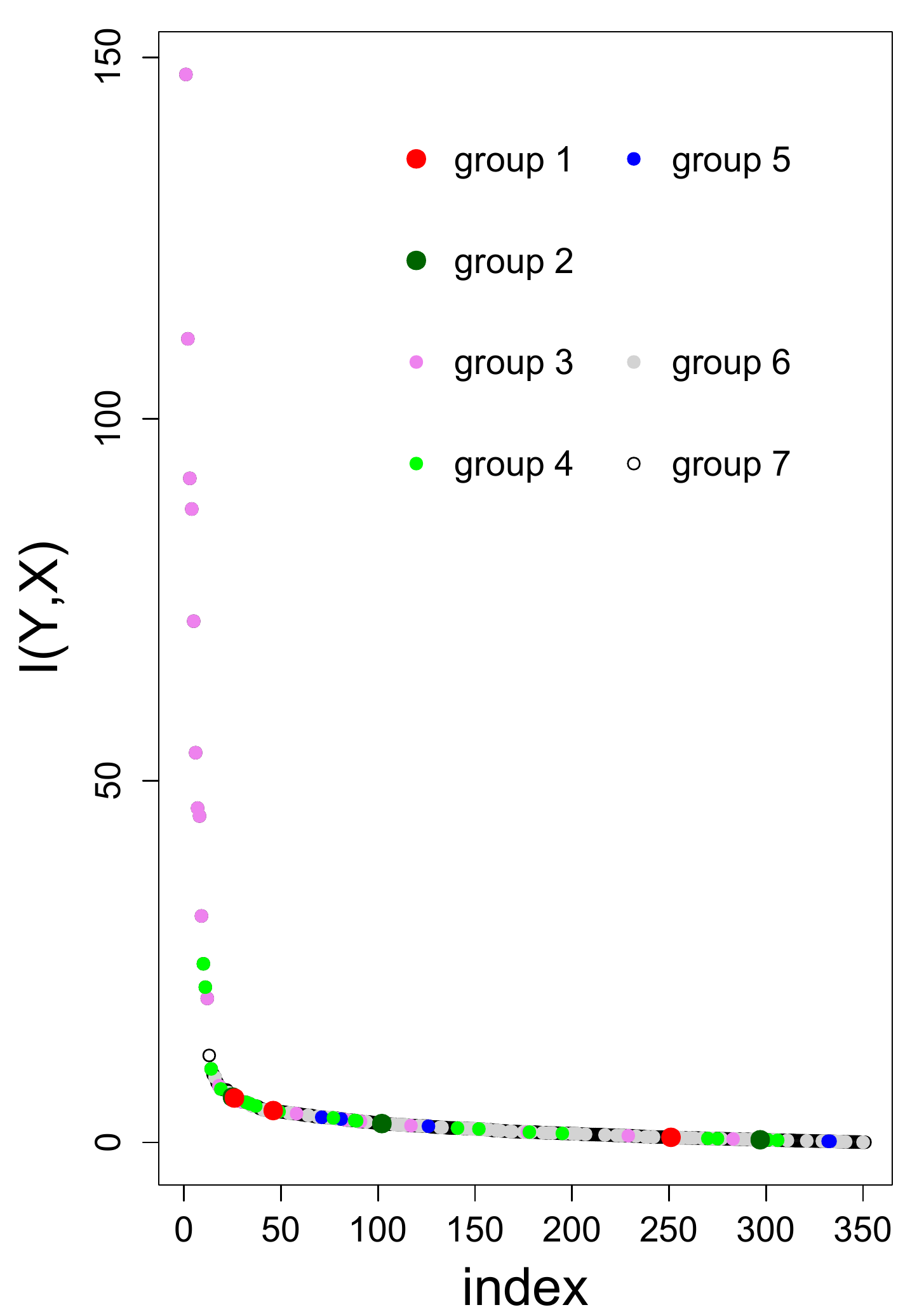}&
\includegraphics[width=0.28\textwidth]{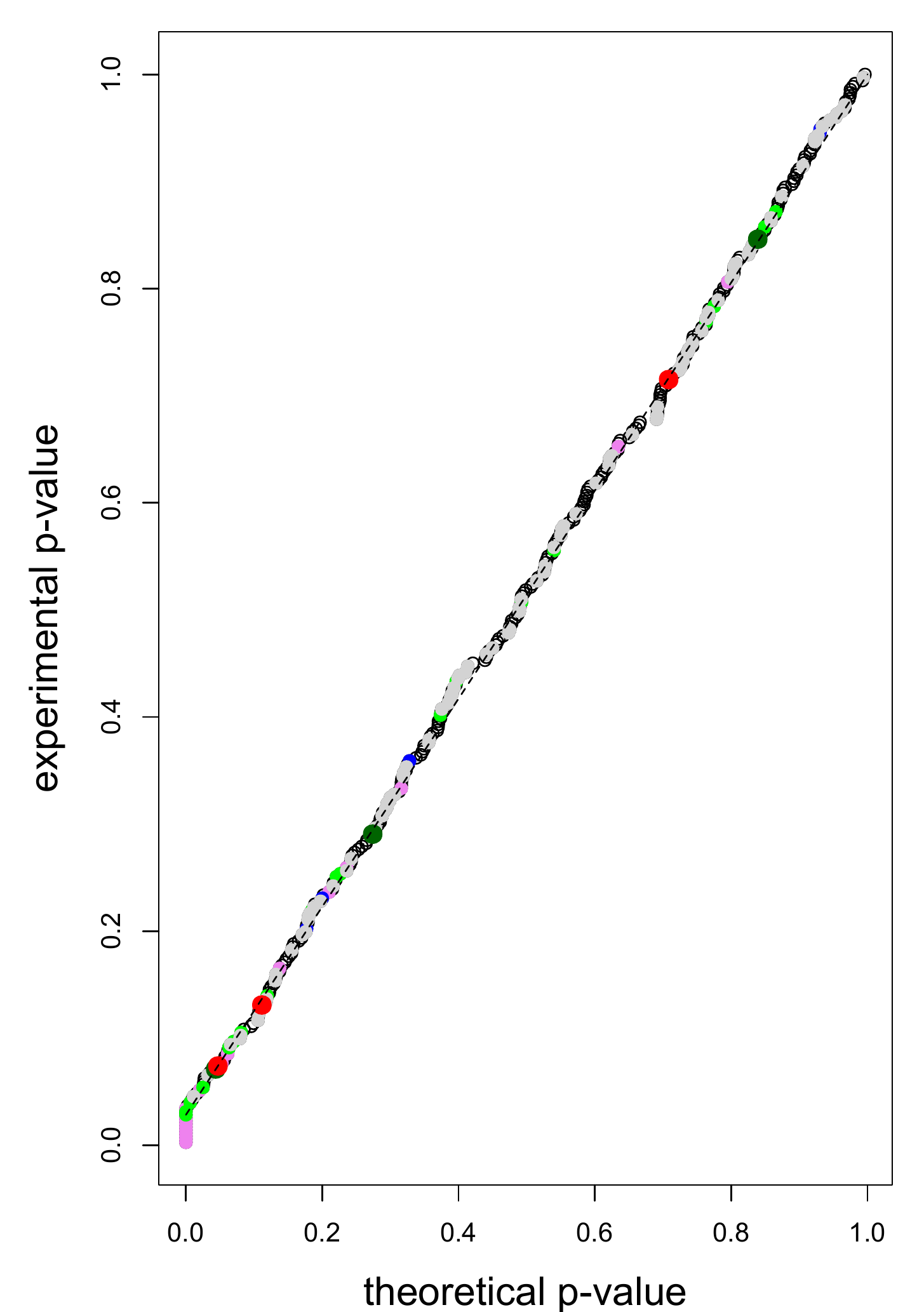}&
\includegraphics[width=0.28\textwidth]{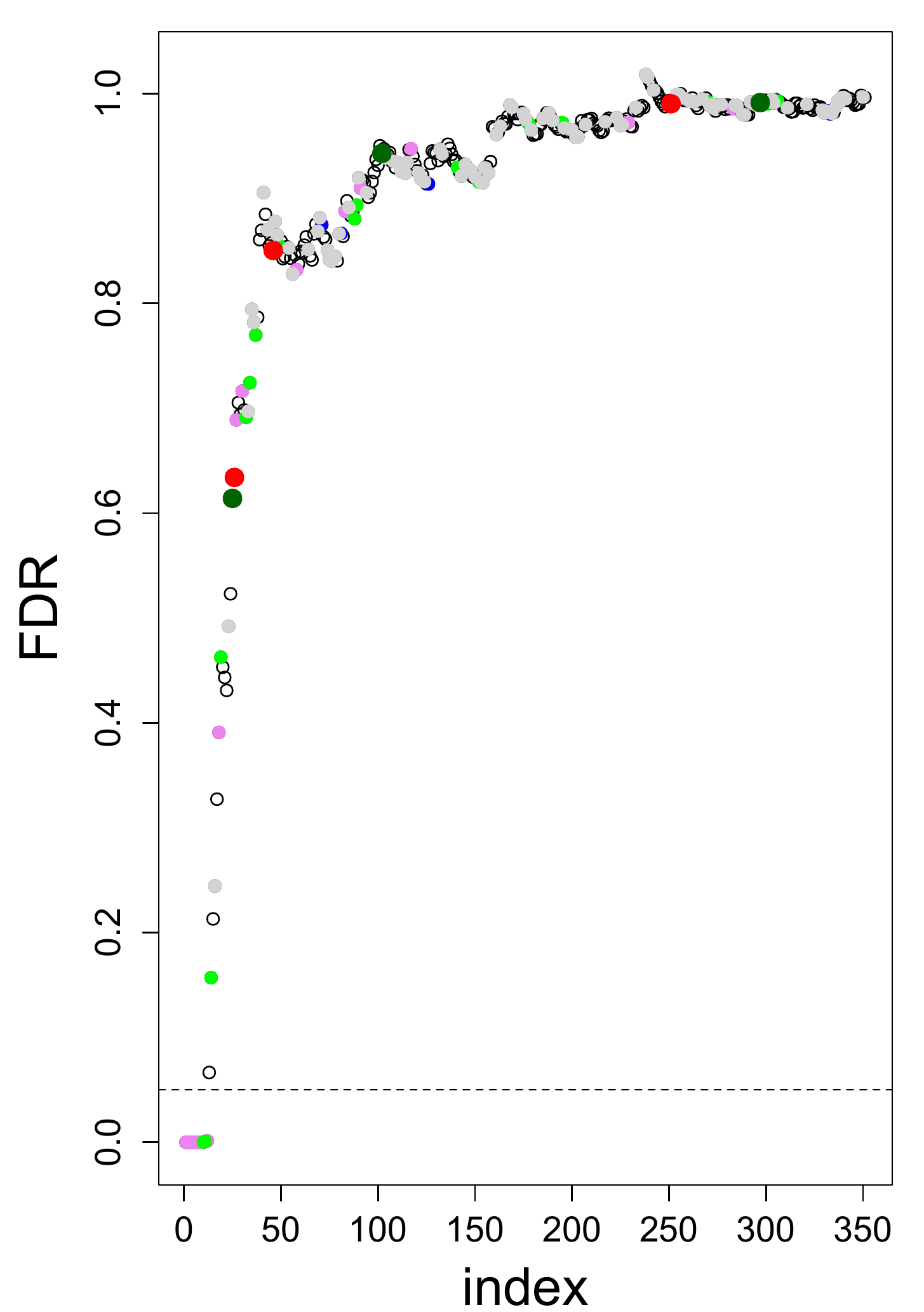}\\
\raisebox{0.2\textwidth}{2D}&
\includegraphics[width=0.28\textwidth]{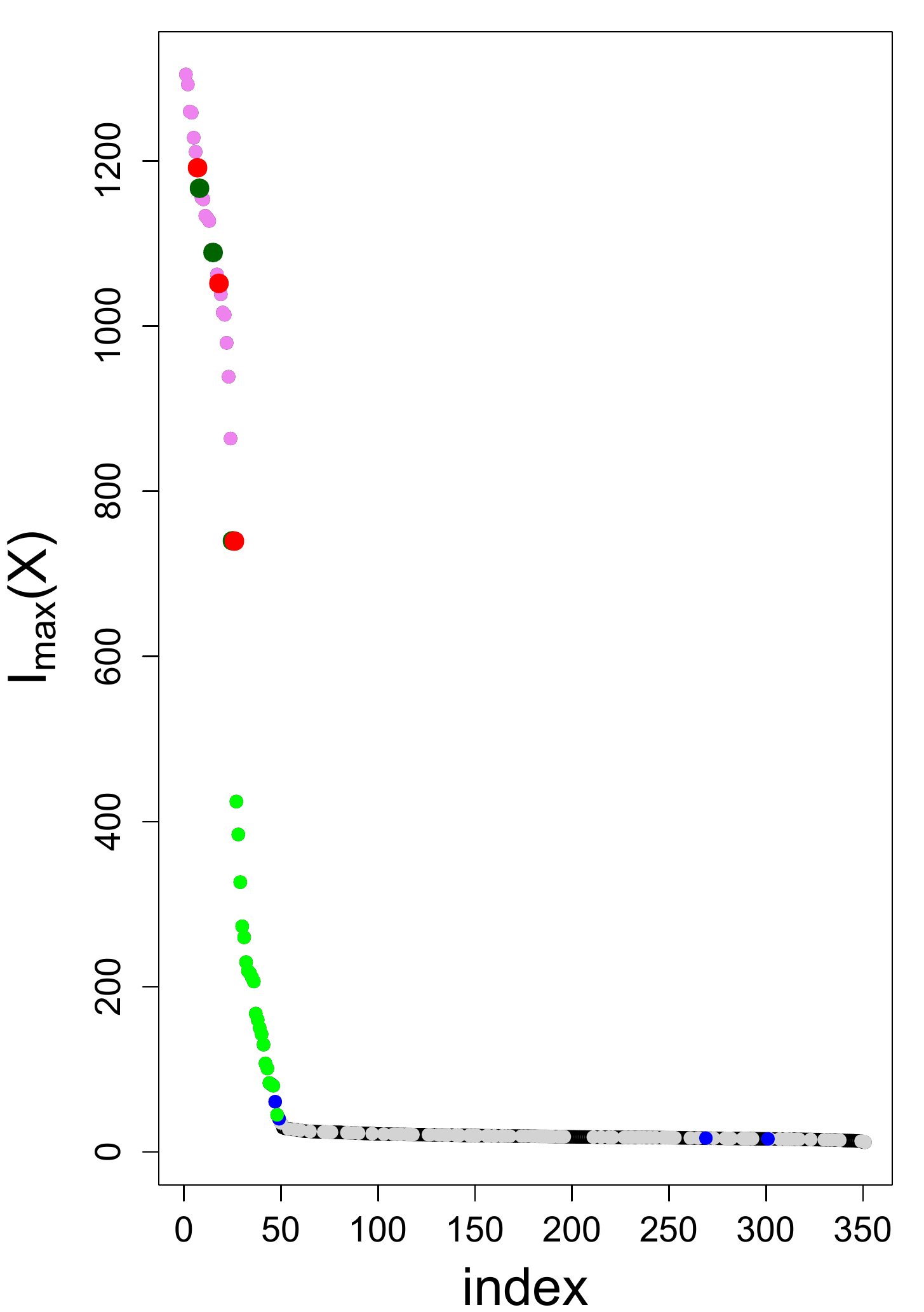}&
\includegraphics[width=0.28\textwidth]{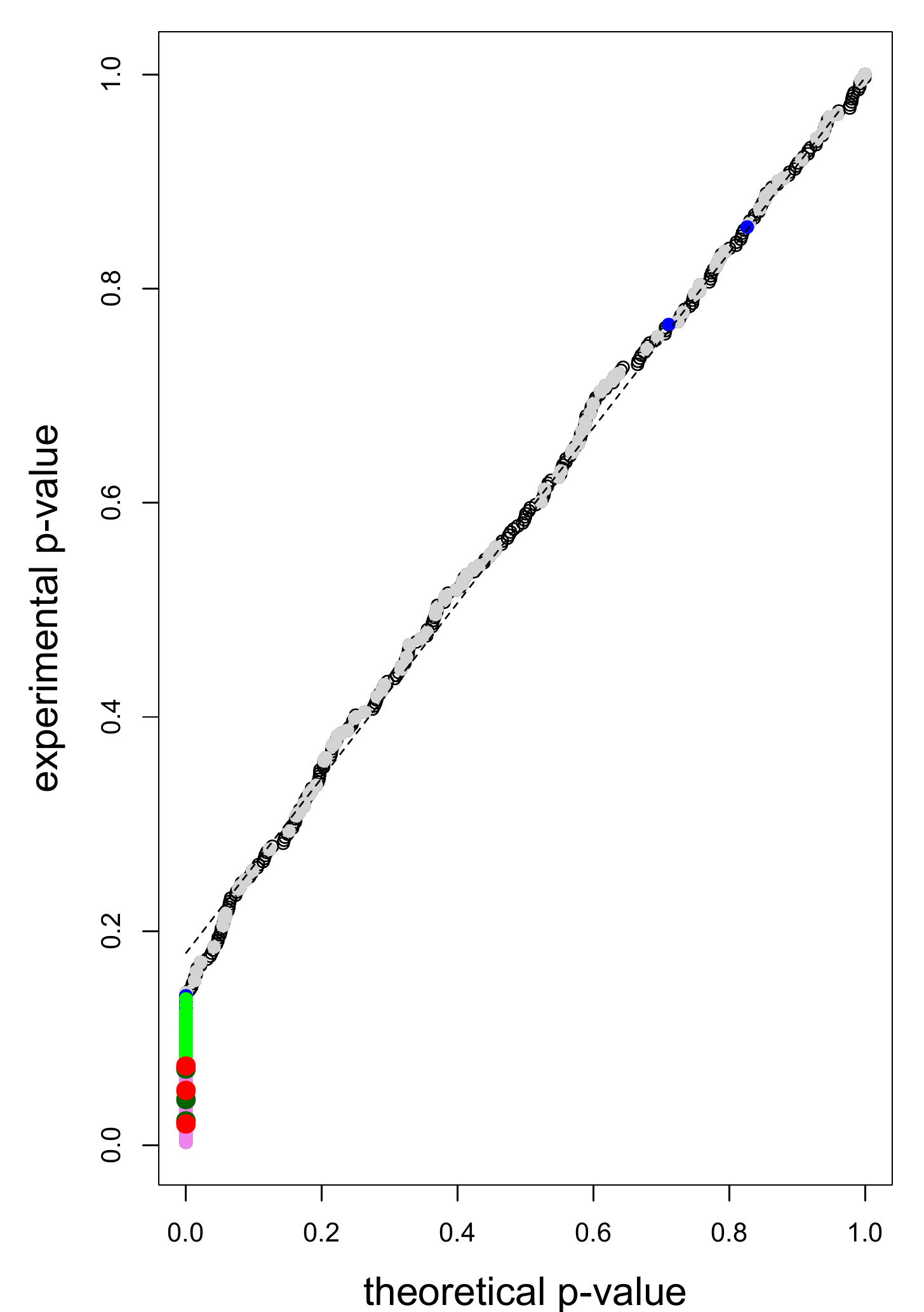}&
\includegraphics[width=0.28\textwidth]{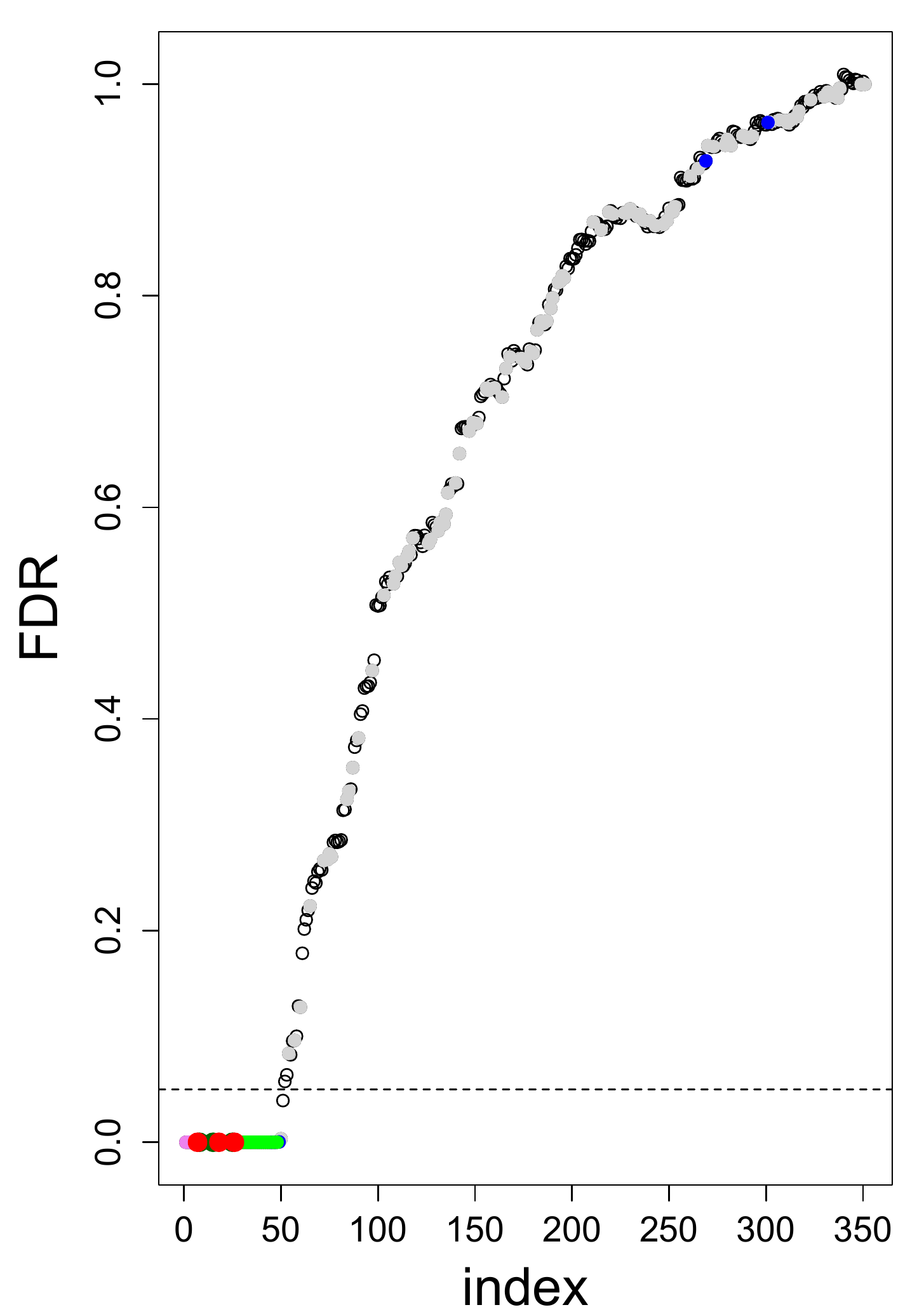}\\
\raisebox{0.2\textwidth}{3D}&
\includegraphics[width=0.28\textwidth]{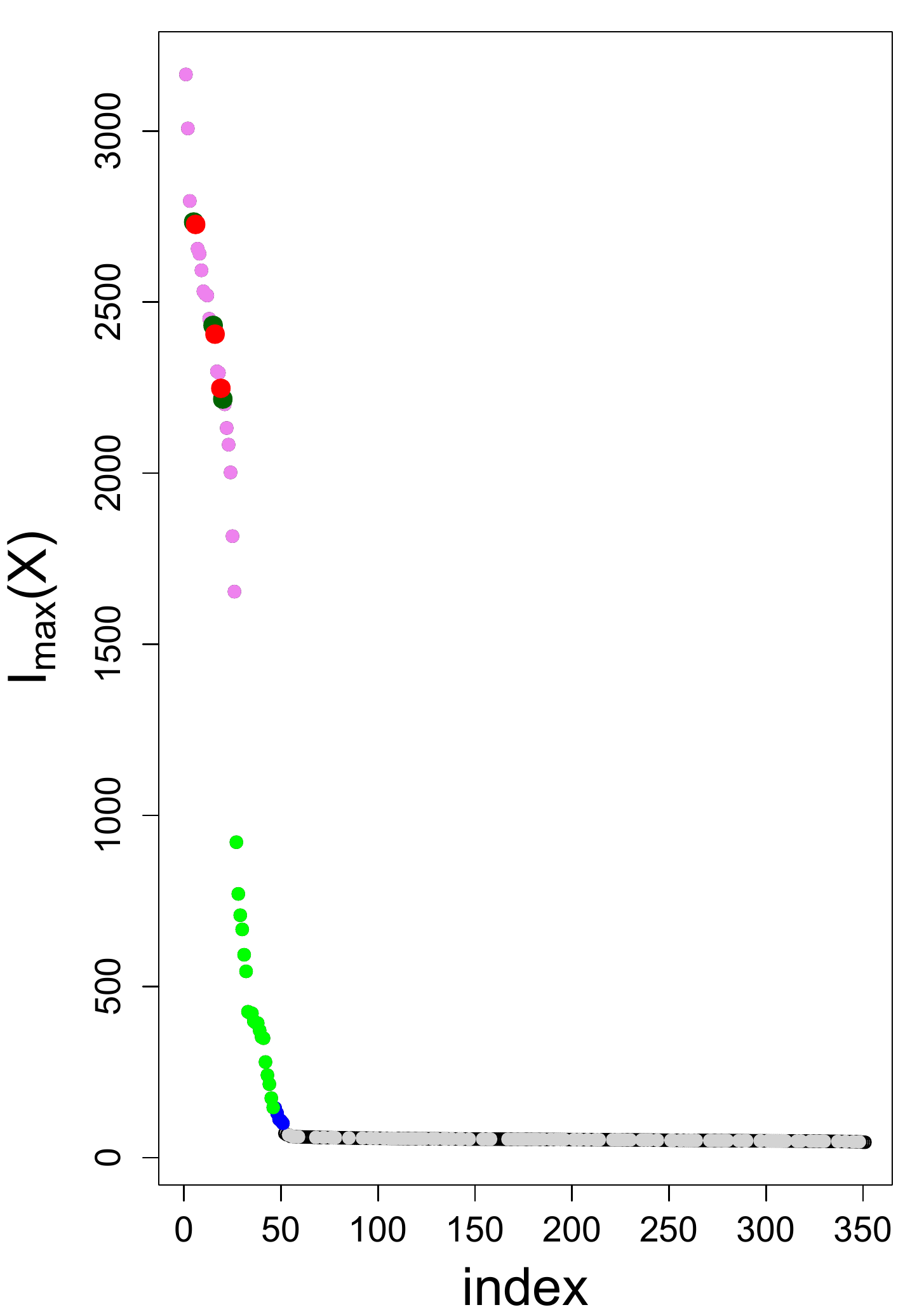}&
\includegraphics[width=0.28\textwidth]{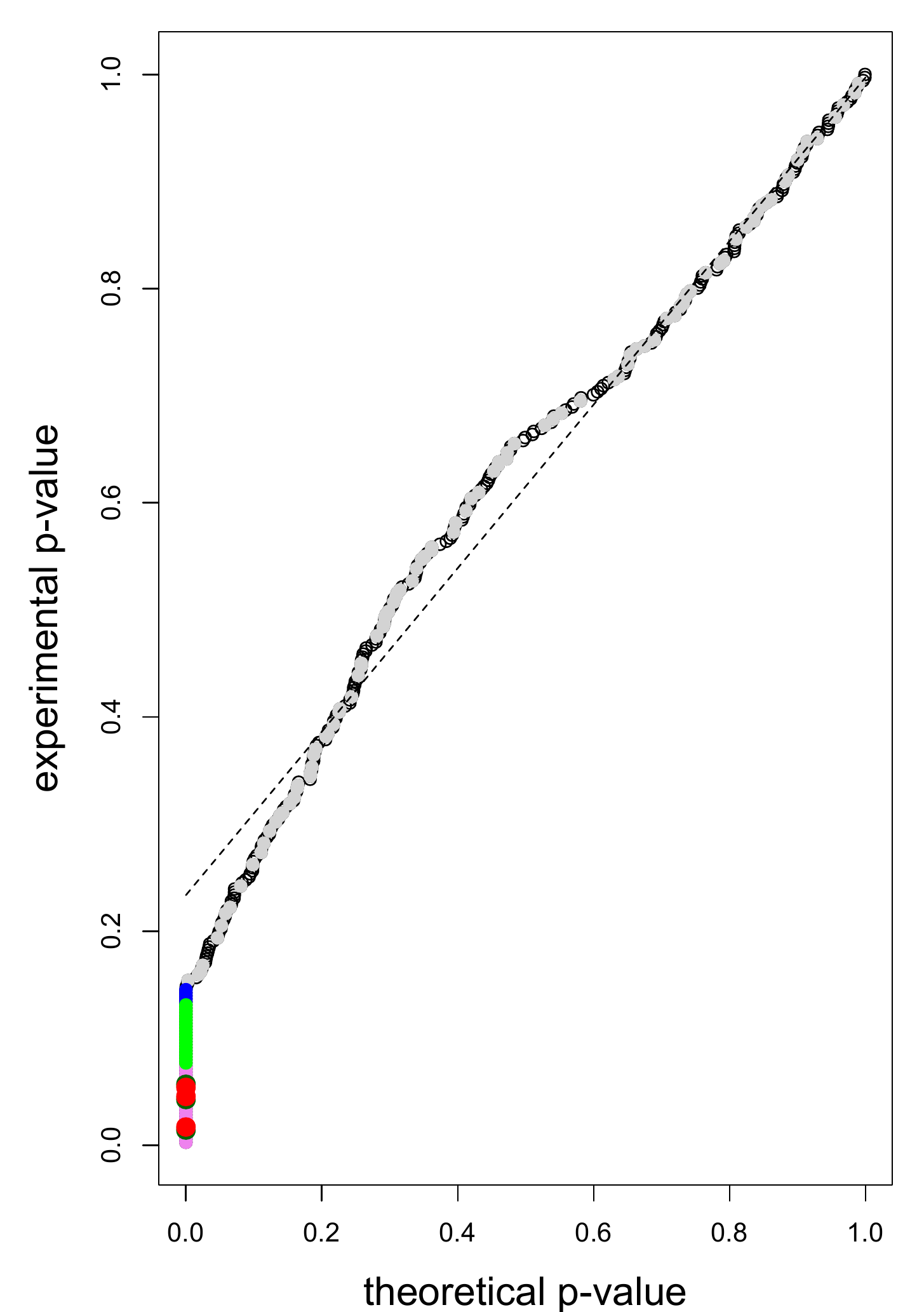}&
\includegraphics[width=0.28\textwidth]{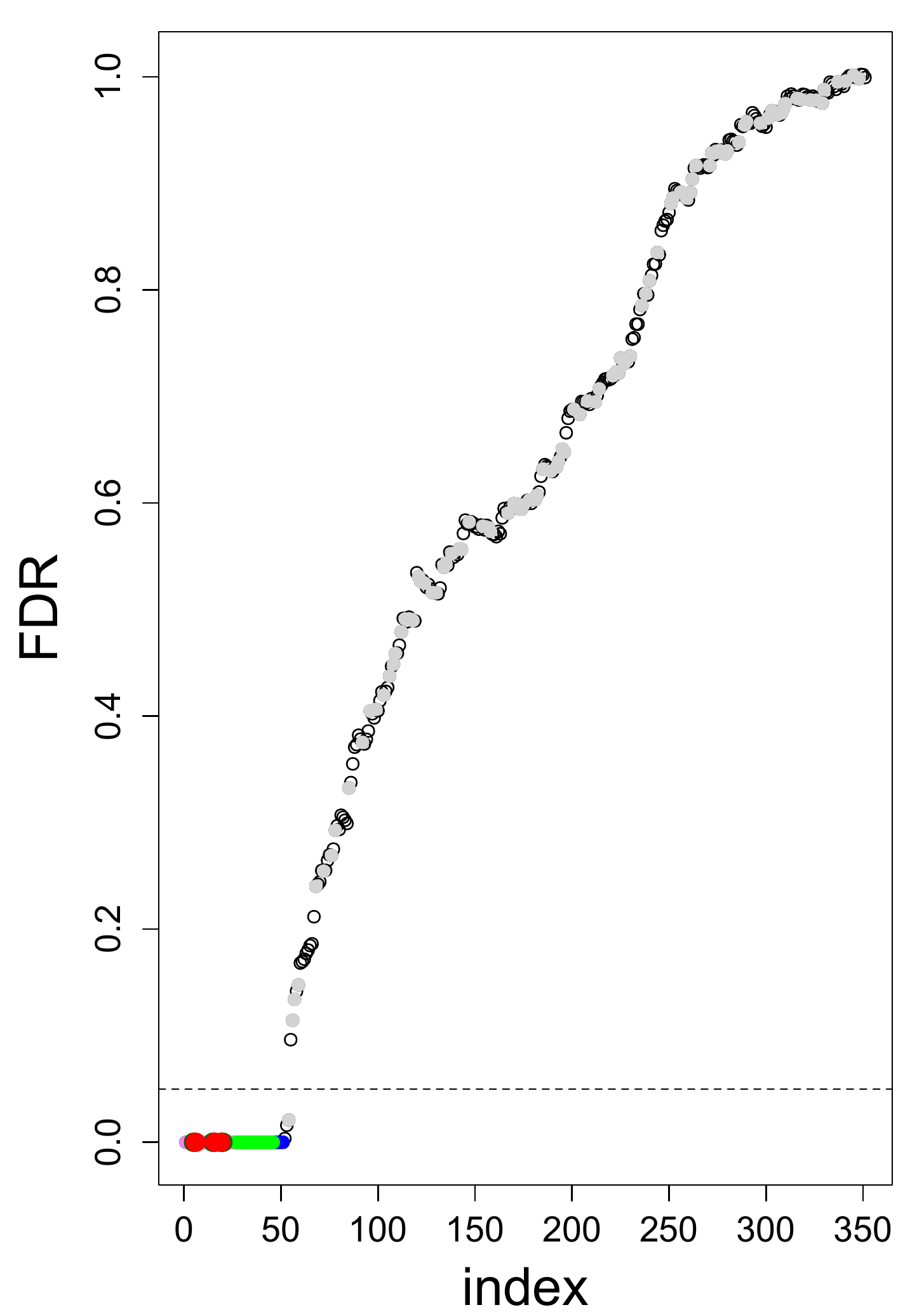}\\
\end{tabular}
\caption{Test results for 1-, 2- and 3-dimensional analysis of relevance for response variable defined as $Y=( X_1\cdot X_2 \cdot X_3 <0 )$ (3-dimensional XOR function). 1st, 2nd, and 3rd row correspond to 1D, 2D and 3D analysis, respectively. 
The plots in the first column show the information gain, in the second one the p-p plot for the theoretical vs. experimental distribution. 
The last column contains plots of the expected value of false discovery rate. 
}
\label{results1}
\end{figure} 

The graphical summary presented in the Figure~\ref{results1} shows the results of three stages for 1-, 2- and 3-dimensional analysis. 
The analysis in 1-, 2- and 3-dimensions are shown in the 1st, 2nd and 3rd row, respectively. 
The first column in the figure shows the gain of information on the decision variable due to knowledge of the variable under scrutiny, for all variables. 
The second column shows the P-P plot for the maximal information gain (2D and 3D case) or standard chi-squared distribution in 1D. 
The last column contains plots of the expected value of false discovery rate computed using Benjamini-Hochberg method.


\begin{figure}
\centering
\begin{tabular}{ccc}
 3D sphere & 3D XOR & 3D sinus \\
\includegraphics[width=0.31\textwidth]{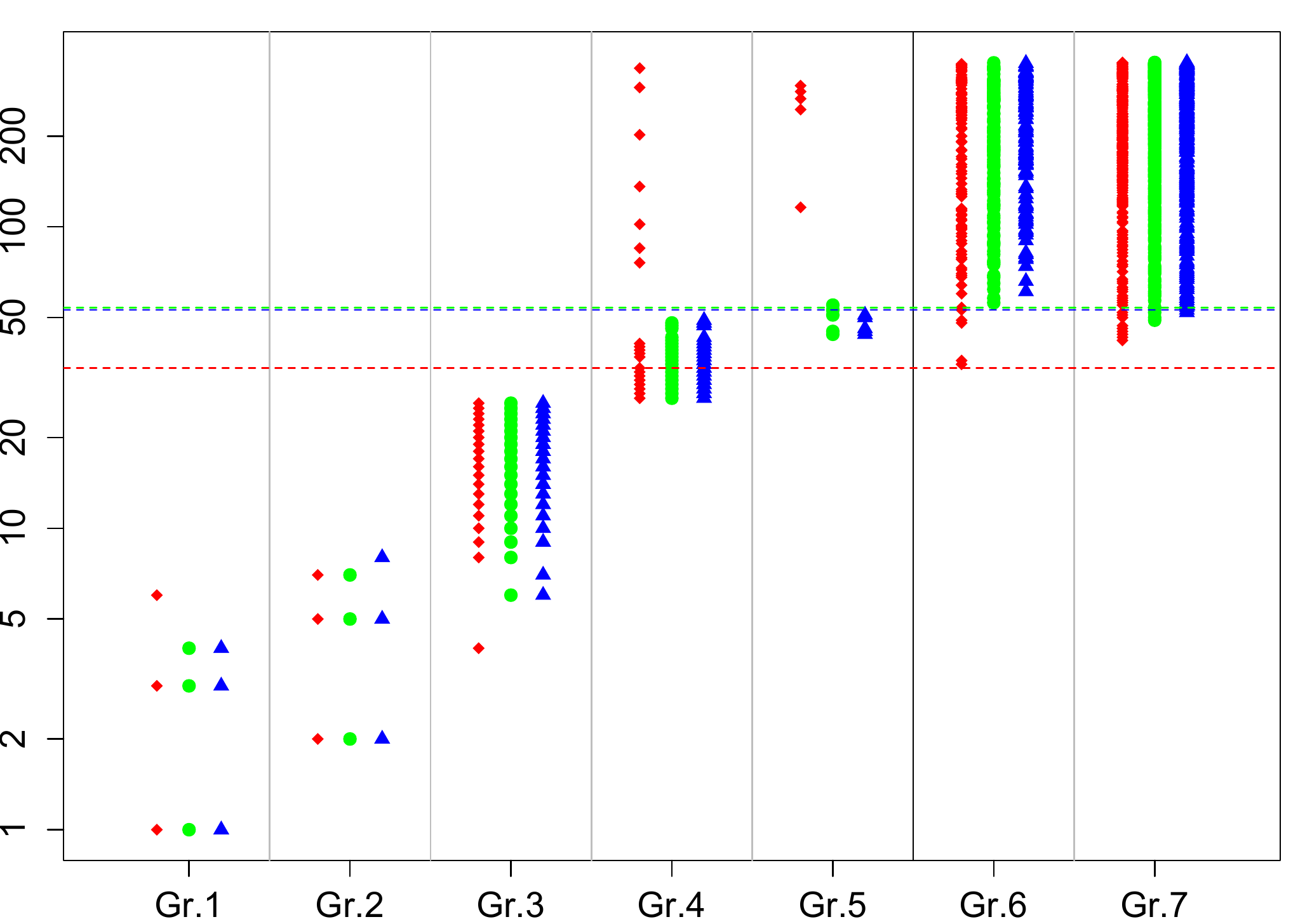}&
\includegraphics[width=0.31\textwidth]{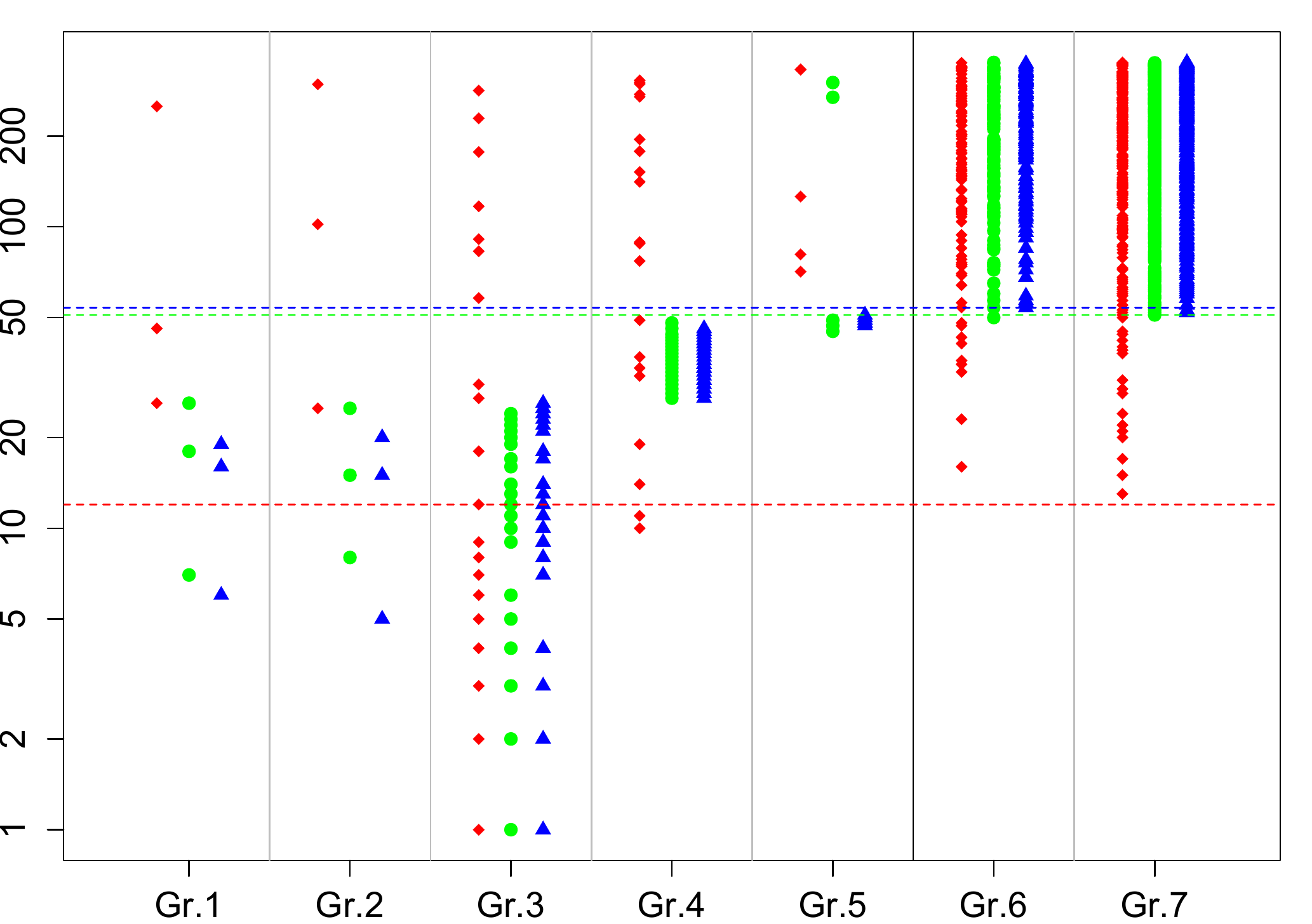}&
\includegraphics[width=0.31\textwidth]{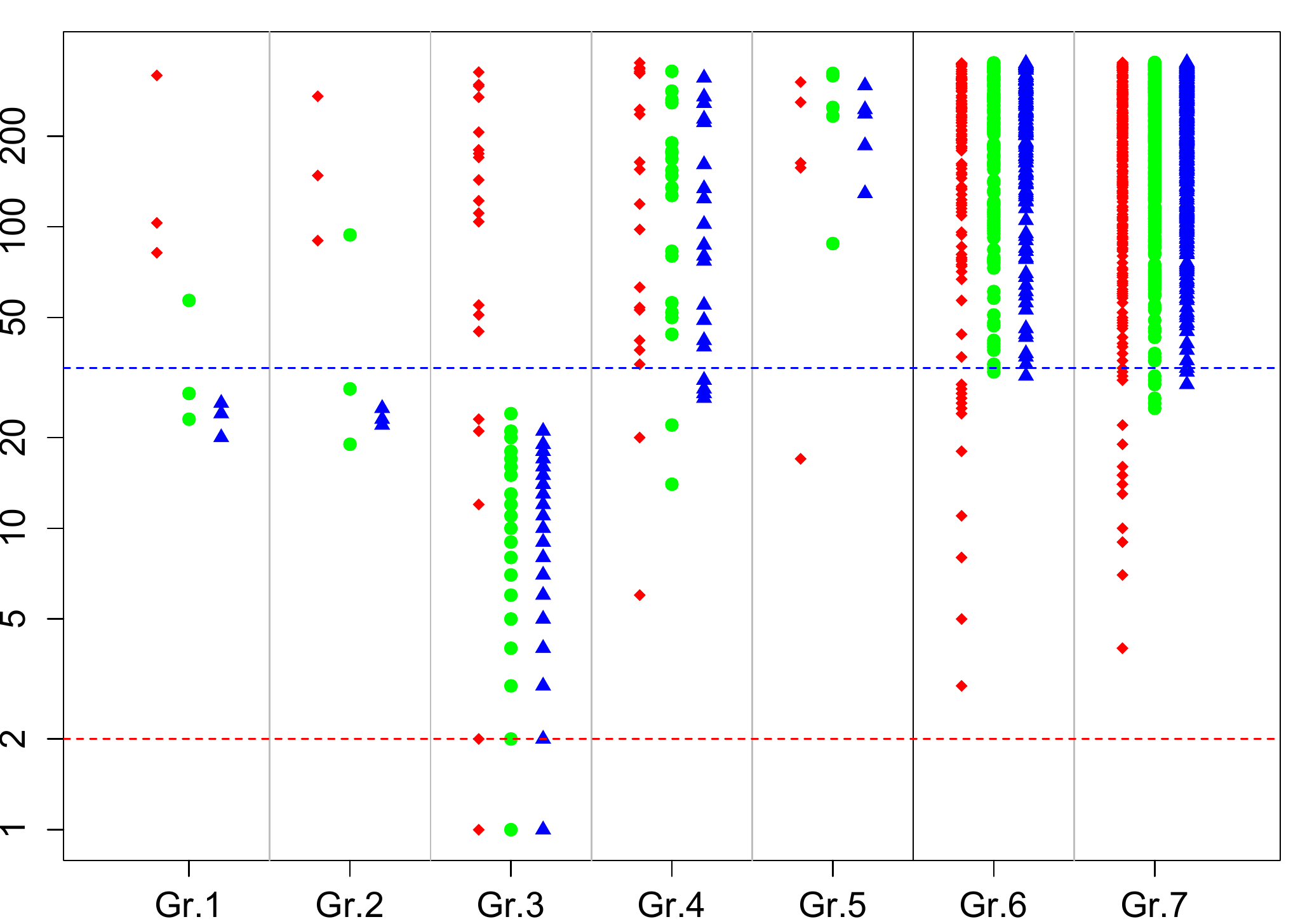}\\
\end{tabular}
\caption{The distribution of ranks for 1-, 2- and 3-dimensional analysis of relevance for three response variables. The ranks for each group of variables are shown side by side for 1D, 2D and 3D analysis.}
\label{ranks_figure}
\end{figure}




Particularly interesting is the excellent concordance between the theoretical and experimental distribution for the irrelevant variables on the P-P plots. 
In 1D p-values are computed for the chi-squared with 2 degrees of freedom, whereas for 2D and 3D they are computed from the extreme value distribution, with the distribution parameter estimated from data. 
The fit is also reasonably good, although not perfect in 3D. 
We suspect that this is due to larger number of bins what can increase variance of the underlying chi-squared distribution.  
The final step of the algorithm, namely calling the relevant variables, is illustrated in the third column of the Figure~\ref{results1}. 
One can observe qualitative difference between the 1D on one side and 2D and 3D on the other. 
In the former, the FDR stays at zero for few variables, and then very rapidly approaches one, whereas for two latter cases the first horizontal segment is much longer and the following increase towards one is much more gradual. 
This reflects the differences between distribution of ranks in these cases, see Figure~\ref{ranks_figure}. 
In the 1D case, only few truly relevant variables score high, whereas the distribution of ranks for the remaining is random.
In 2D and 3D, nearly all relevant variables are ranked higher than the irrelevant ones. 

\subsection{Hard problem - 3D checkerboard}

\begin{figure}
\centering
\begin{tabular}{ccc}
$I_{max}(X)$ & $p$-value & FDR \\
\includegraphics[width=0.31\textwidth]{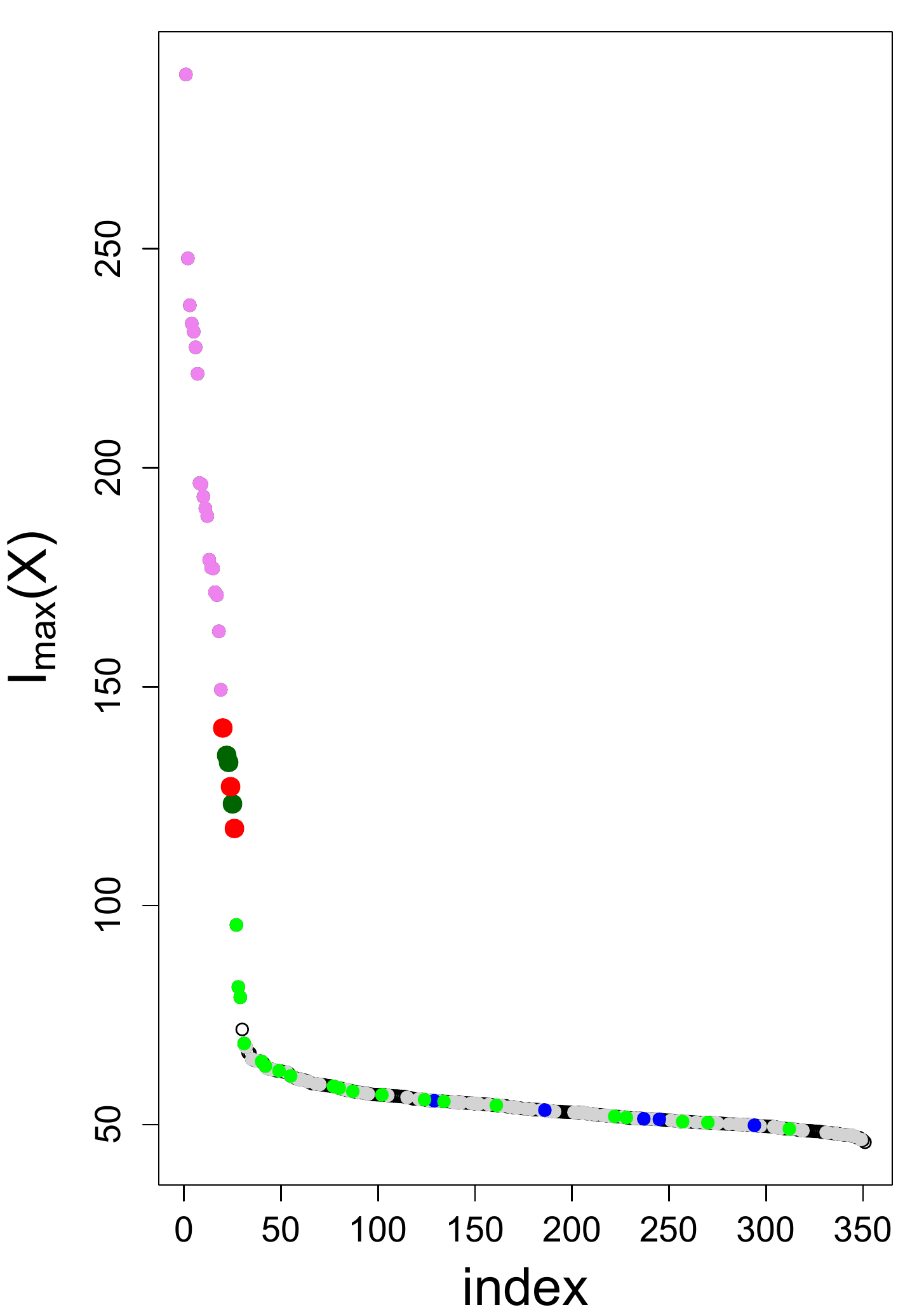}&
\includegraphics[width=0.31\textwidth]{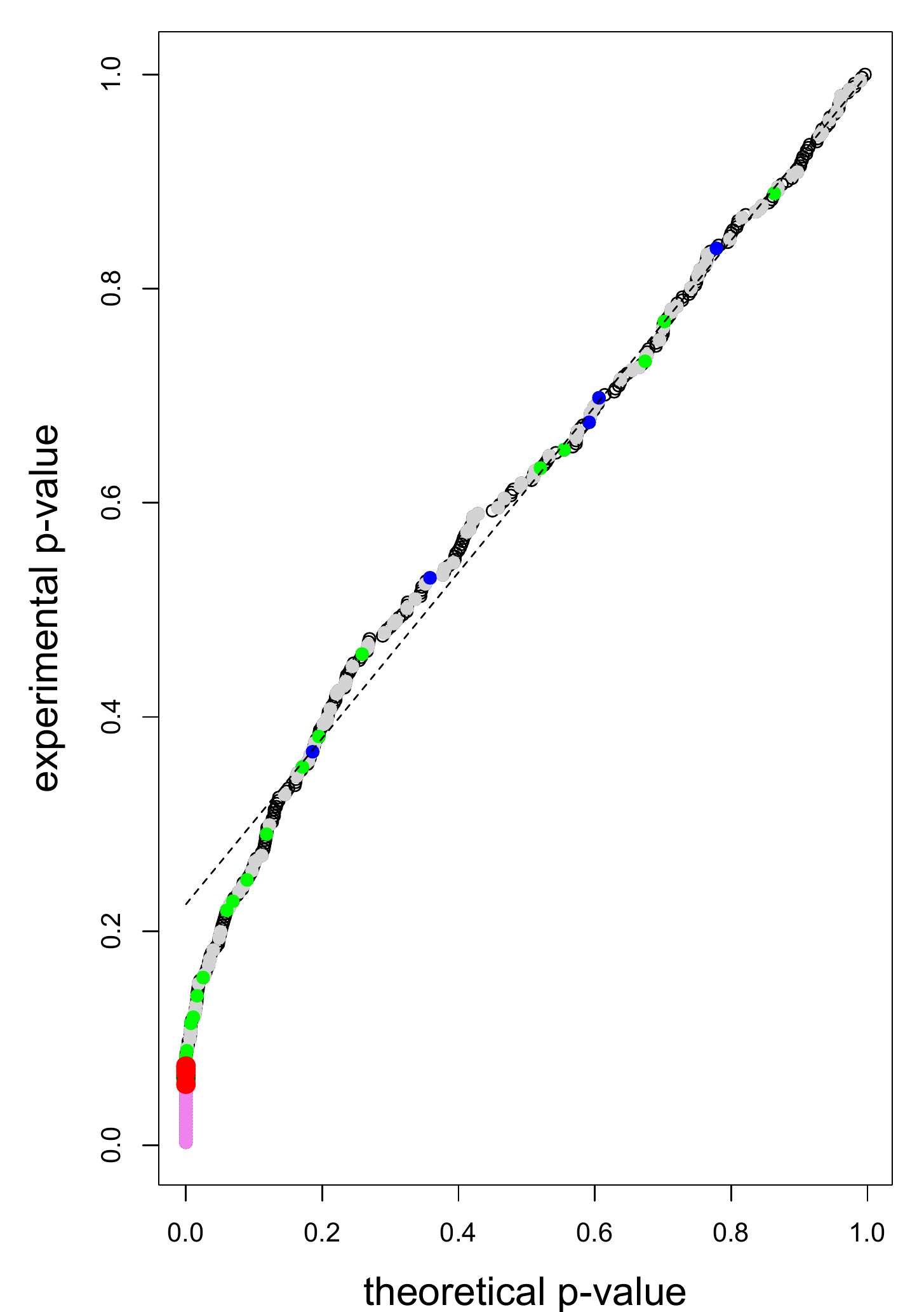}&
\includegraphics[width=0.31\textwidth]{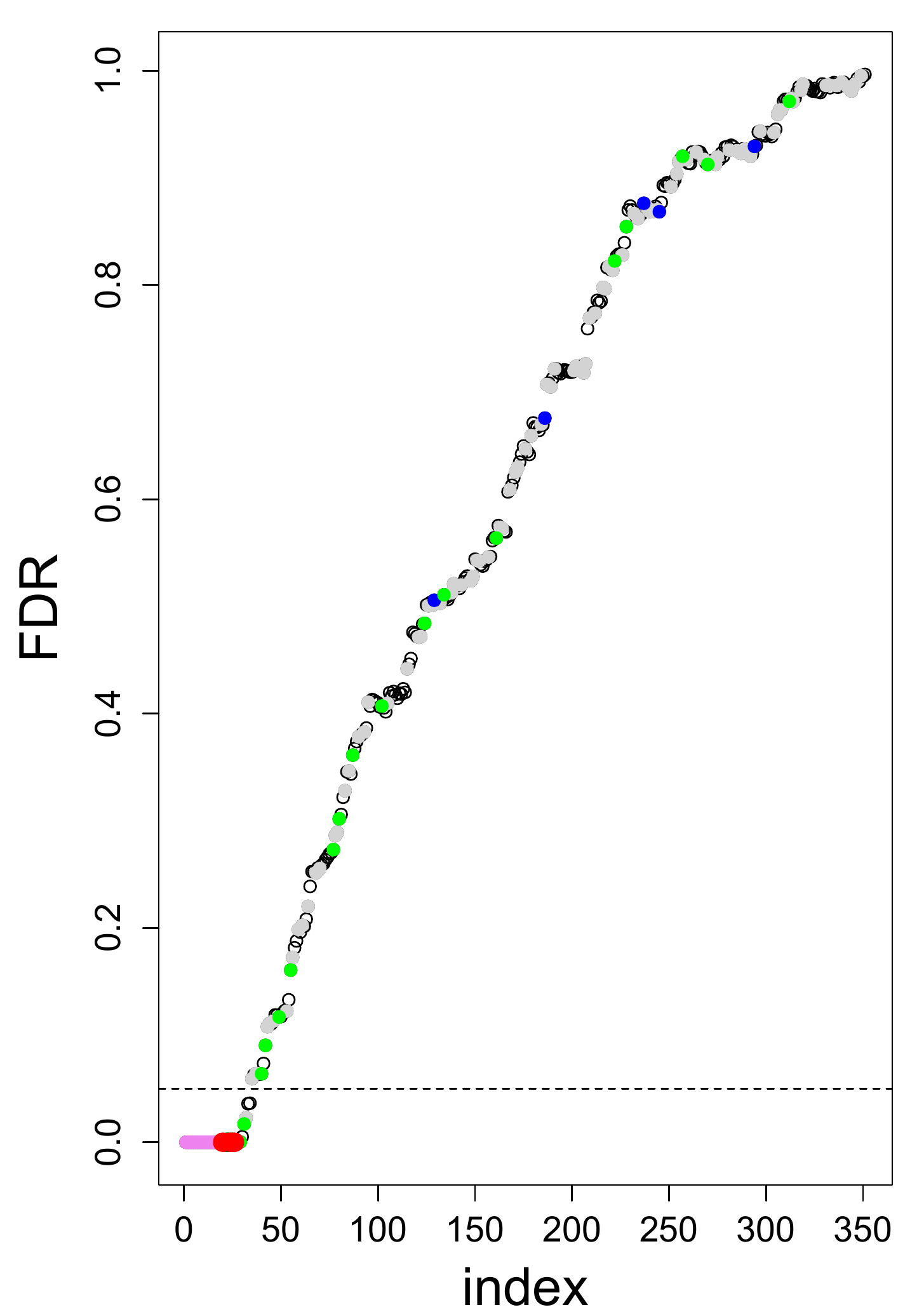}\\
\end{tabular}
\caption{Test results for 3-dimensional analysis of relevance for response variable defined as $Y=( \sin(2 \pi X_1) \cdot \sin(2 \pi X_1) \cdot \sin(2 \pi X_1) <0 )$ (3-dimensional checkerboard pattern). See Fig.~\ref{results1} for detailed description.}
\label{results_sin}
\end{figure}

The third example of the informative response function generated using the same descriptor set is a 3-dimensional product of sinus functions that generates 3-dimensional $4\times4\times4$ checkerboard-like pattern. 
This function is another example of pure synergy when analysed for base descriptors only, with more complex distribution of the response function.  
What is more, three equipotent splits of descriptors don't align well with variation of the decision function what makes the analysis harder. 
This mismatch results in much smaller information gains attributed to informative variables than in the previous example, and lower sensitivity, in particular in lower dimensions. 
Only two linear combinations of base variables were detected as relevant in 1-dimensional analysis.  
The sensitivity of the 2-dimensional analysis is higher, resulting in discovery of all of the pure linear combinations along with two of base variables. 
Finally, the 3-dimensional analysis revealed all the base variables, both pure and distorted,  all pure linear combinations and some of the noisy linear combinations, see Figure~\ref{results_sin}. 
Furthermore, that even in 3D base variables were ranked lower than all but one linear combinations, see Figure~\ref{ranks_figure}. 
The analysis based on highest scoring variables could miss these most important variables. 

\subsection{Random decision function}
The final test is performed on the random response variable that is not dependent on any descriptors. 
This test is performed to check whether the procedure is not too aggressive and able to call random variables as relevant if there are no true relevant variables. 
The test is passed, since no variable has been called relevant neither in 1D, nor 2D nor 3D analysis. 
Additionally, the rankings of variables are completely random. 

\subsection{Classification using identified relevant variables}
The results of feature selection were further used for building predictive Random Forest models to test whether by extending the search for informative variables into higher dimension can potentially improve the modelling results. 
In particular, we were interested in the problem, where, on the one hand a significant fraction of relevant variables were not discovered using 1-dimensional analysis, but on the other hand several relevant variables were found. 
Therefore, we concentrated on the 3D XOR, since two other problems are either too easy (3D sphere) or too hard (3D checkerboard). 

It turns out, that the 1-dimensional analysis reveals too few variables for building high quality models, either using all or 3-best variables only, see Table~\ref{OOB_error}. 
However, both 2- and 3-dimensional analysis find subset of variables sufficient to build high quality models, with quality comparable to those built using all informative variables.  
Interestingly, the models build on three highest scoring variables are significantly worse than models based on all relevant variables, with the error level roughly six times higher. 
It is worthwile to note, that the models built using just three base variables are significantly better than models built using all informative variables. 
This last observation shows that in some cases identification of the causal variables is possible. 

The results of this analysis show that in the presence of synergistic interactions between variables the extension of the analysis into multiple dimensions allows to find variables involved in these interactions and to improve predictive models. 

\begin{table}
\caption{OOB classification error of the Random Forest classifier built using features selected using 1-,2- and 3-dimensional analysis on the XOR dataset, in comparison with results obtained by using predefines subsets of variables: B - base, LC - linear combinations, B + LC - base plus linear combinations, and AR - all relevant variables. }
\begin{tabular}{r | ccc || cccc}
					&		&		&		& \multicolumn{4}{c}{Reference} \\
Selected variables	&	1D 	&	2D 	&	3D	& 	B	& B + LC&	LC	&	AR	\\
\hline
All 				& 5.4\%	& 1.9\%	& 1.7\%	& 0.4\%	& 1.4\%	& 3.1\%	& 1.5\%	\\
3 top ranking 		& 9.4\%	& 8.9\%	& 11.4\%& 		& 		& 		& 		\\
\end{tabular}
\label{OOB_error}
\end{table}

\begin{figure}
\centering
\begin{tabular}{ccc}
$I_{max}(X)$ & $p$-value & FDR \\
\includegraphics[width=0.28\textwidth]{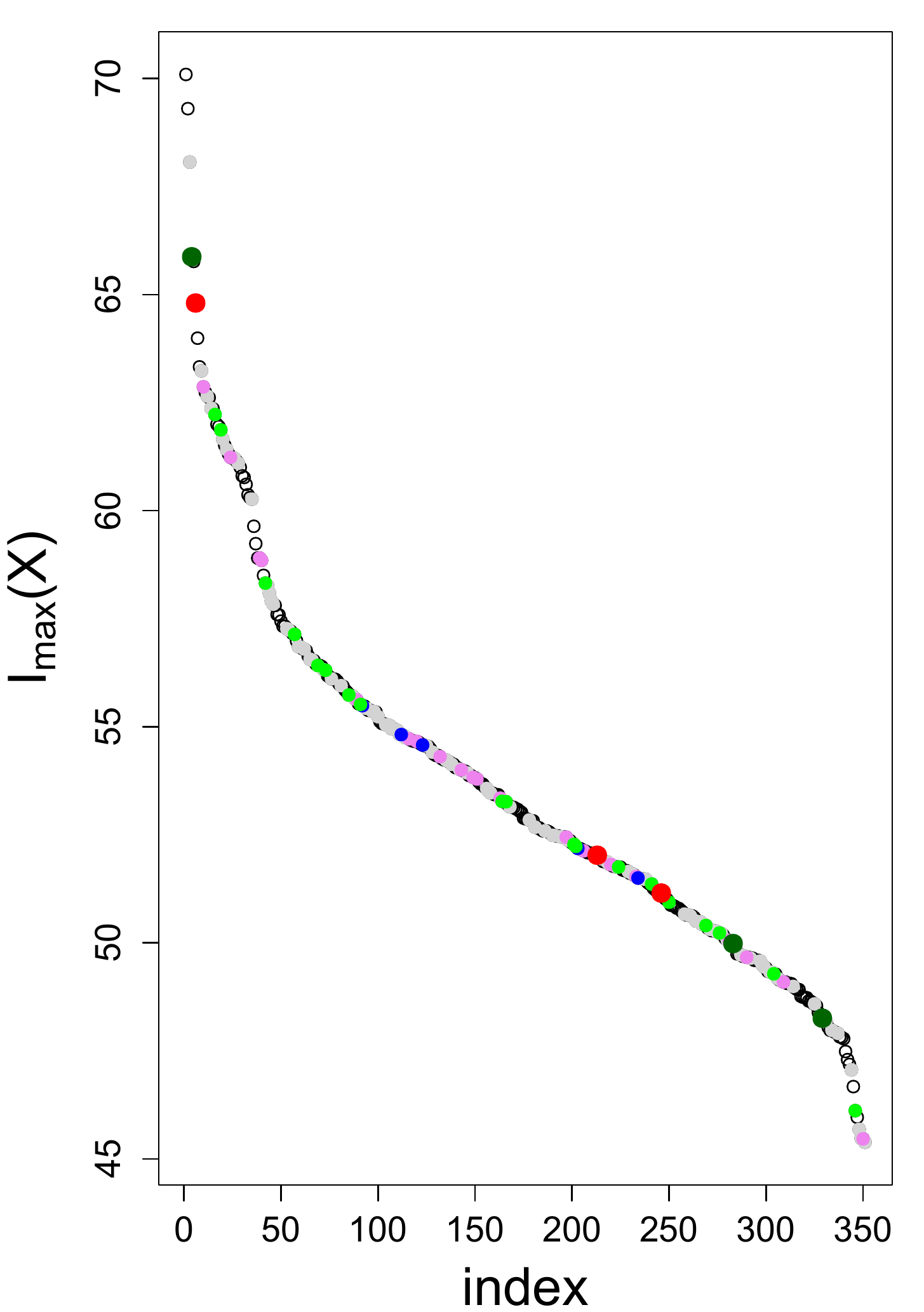}&
\includegraphics[width=0.28\textwidth]{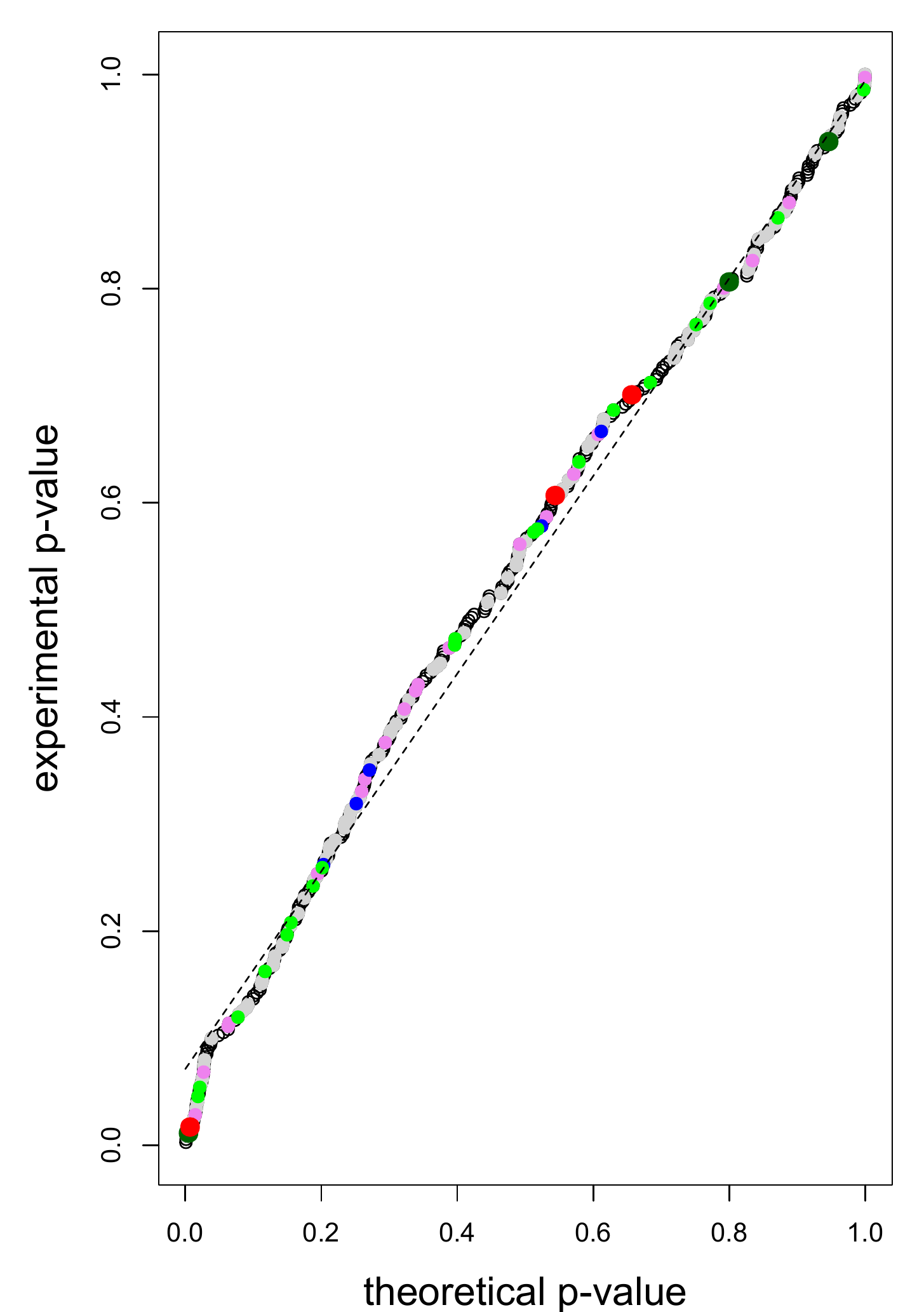}&
\includegraphics[width=0.28\textwidth]{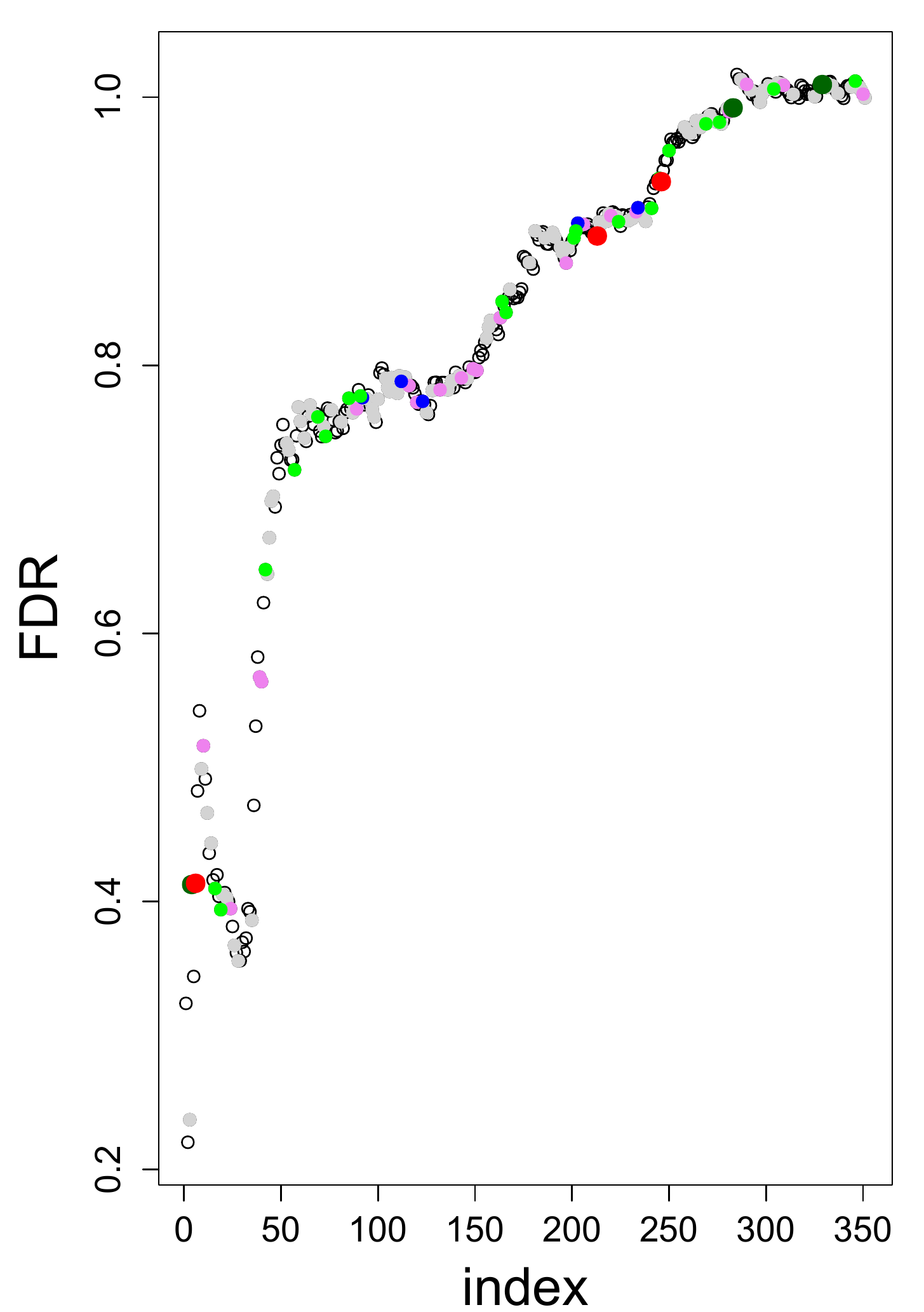}\\
\end{tabular}
\caption{Test results for 3-dimensional analysis of relevance for random response variable. See Fig.~\ref{results1} for detailed description.}
\label{results4}
\end{figure}

%% file: fs6_conclusions.tex
\section{Conclusions}

In the current study we have explored the possibility of identification of relevant variables, when the response variable is a complex multidimensional function of the descriptive features with nonlinear and synergistic interactions between variables.  
To this end, we have proposed a rigorous methodology for identification of all the relevant variables by the exhaustive multidimensional analysis using information theory. 
Based on the solid statistics, the method allows either for the false discovery control using Benjamini–Hochberg procedure, or family-wise error rate, for example using Bonferoni correction. 
We have tested the methodology using the complex multidimensional datasets where response variable was a 3-dimensional nonlinear function of 3 descriptors and descriptor set contained also linear combinations of base variables, noisy copies of relevant variables noisy combinations of base variables, nuisance variables as well as the random noise. 
We have demonstrated that univariate analysis fails to identify most of the relevant variables in truly multidimensional cases and requires multidimensional analysis.  

The tests on artificial data confirmed that the new approach is able to identify all or nearly all the weakly relevant variables (in the sense of 
\citec{Kohavi-John}{kohavi_john}), when the dimensionality of the analysis matches the true dimensionality of the problem.  
The exhaustive search of all possible combinations of variables limits the practical dimensionality of the analysis due to the quick growth of the number of combinations with increasing dimensionality. 
Nevertheless, the presence of linear combinations of variables effectively lowers the dimensionality of analysis required for discovering the truly important variables. 
In the case of 3D sphere and 3D XOR, all base variables and all combination variables, as well as most of the nuisance variables were identified already in the 2-dimensional analysis.  
Only the discovery of all subtle nuisance effects required full 3-dimensional analysis concordant with the true dimensionality of the problem. 

Interestingly, in the two more difficult problems the highest scoring variables are not those that were used for the generation of the response variable, independent of the dimensionality of the analysis.  
What is more, in some cases the base variables scored significantly below their linear combinations. 
This result suggests that when the goal of the feature selection is identification of causal variables, then the all-relevant feature selection approach should be applied. 

{\bf Acknowledgement}

The research was funded by the by the Polish National Science Centre, grant DEC-2013/09/B/ST6/01550.

%% file: appendix_a.tex
\subsection{The probability distribution of the conditional mutual information}

\label{df_cond}

The estimate of the conditional mutual information (\ref{estcond}) can be rewritten in the form:
\begin{equation}
  \hat{I}(Y;X|S)
   = \frac{1}{N}\sum_{y,x,\{s_i\}}n_{yx\{s_i\}}\log \frac{\hat{p}_{yx\{s_i\}}\hat{p}_{\{s_i\}}}{\hat{p}_{x\{s_i\}}\hat{p}_{y\{s_i\}}}
\end{equation}
which allow us to identify the expression:
\[\hat{p}^0_{yx\{s_i\}}=\frac{\hat{p}_{y\{s_i\}}\hat{p}_{x\{s_i\}}}{\hat{p}_{\{s_i\}}}\]
as a probability estimate under the null hypothesis. The null hypothesis states, that the variable $X$ contributes no information about $Y$ to the subset $S$. $\hat{p}^0_{yx\{s_i\}}$ is a well-defined probability estimate, since 
\[\sum_{y,x,\{s_i\}}\hat{p}^0_{yx\{s_i\}}=1\]
Hence, $N I(Y;X|S)$ can be interpreted as a likelihood ratio between the null-hypothesis distribution and the estimate distribution of $Y$,~$X$,~$S$. It has been proved by 
\citec{Wilks}{wilks1938}, that 
\[2 \sum_i n_i \log \frac{\hat{p}_i}{\hat{p}^0_i}\]
(where natural logarithm is used), 
follows the $\chi^2$ distribution for $N\longrightarrow\infty$. 
In practice, the distribution is close to the asymptotic if the contingency table is not sparse, i.e. there are some objects for each combination of values of the variables. 


The number of degrees of freedom of the $\chi^2$ distribution can be calculated as 
\begin{equation}
df=df_{est}-df_0
\end{equation}
where $df_{est}$ and $df_0$ denote the number of degrees of freedom for the estimated model and the 
null-hypothesis model, respectively. 
The $df$ is computed using a number of classes of the response variable, 
the variable under scrutiny and of all variables in the subset $S_i$ using the following formula:\\ 

\begin{equation}
df 
  =\left( C_Y-1 \right) \left( C_X-1 \right)  \prod_i C_{S_i}
\end{equation}

\subsection{The statistics of $p_{min}(X)$}
\label{statmax}

For each variable $X$ multiple statistic tests are performed, over all possible subsets $S^{k-1}$. 
For these variables that contribute no information about the response variable $Y$, each test leads to a $\chi^2$-distributed quantity.

If the tests were mutually independent, then the minimum $p$-value over all the tests would follow the distribution:
\begin{equation}
P(p_{min}(X)<v)=1-(1-v)^n\approx 1-e^{-nv}
\end{equation}
for a sufficiently big number of tests $n$. 

In our case, the tests are {\em not} independent -- especially for dimensions higher than 2, where many subsets contain common variables. 
However, the number of tests is large enough to produce the exponential distribution, although the value of the distribution parameter is not known a priori; it can be significantly smaller than the number of tests.
\begin{equation}
P(p_{min}(X)<v)\approx 1-e^{-\gamma v},\; \gamma<n
\end{equation}
The distribution parameter $\gamma$ can be estimated a posteriori. 
Assuming that the investigated dataset contains non-negligible number of irrelevant variables. 
For those variables the $p_{min}$  follows the same exponential distribution end they can be used for the estimation of the distribution parameter.
The number of irrelevant variables should be sufficient for that in most cases. 
If it is not the case, one can extend the dataset using the non-informative contrast variables  \citep{Stoppiglia2003,Kursa2010a}. 

The maximum likelihood estimate of $\gamma$ is the reciprocal of the average $p_{min}$:
\[\hat{\gamma}=\frac{1}{<p_{min}>}\]
The average should be taken over the variables, that fit the exponential distribution -- the exceptionally small values of $p_{min}$ are supposed to correspond to relevant variables and should be ignored. 

\begin{figure}[t]
\centering
\begin{tabular}{ll}
{\bf a)}&{\bf b)}\\
\includegraphics[width=0.48\textwidth]{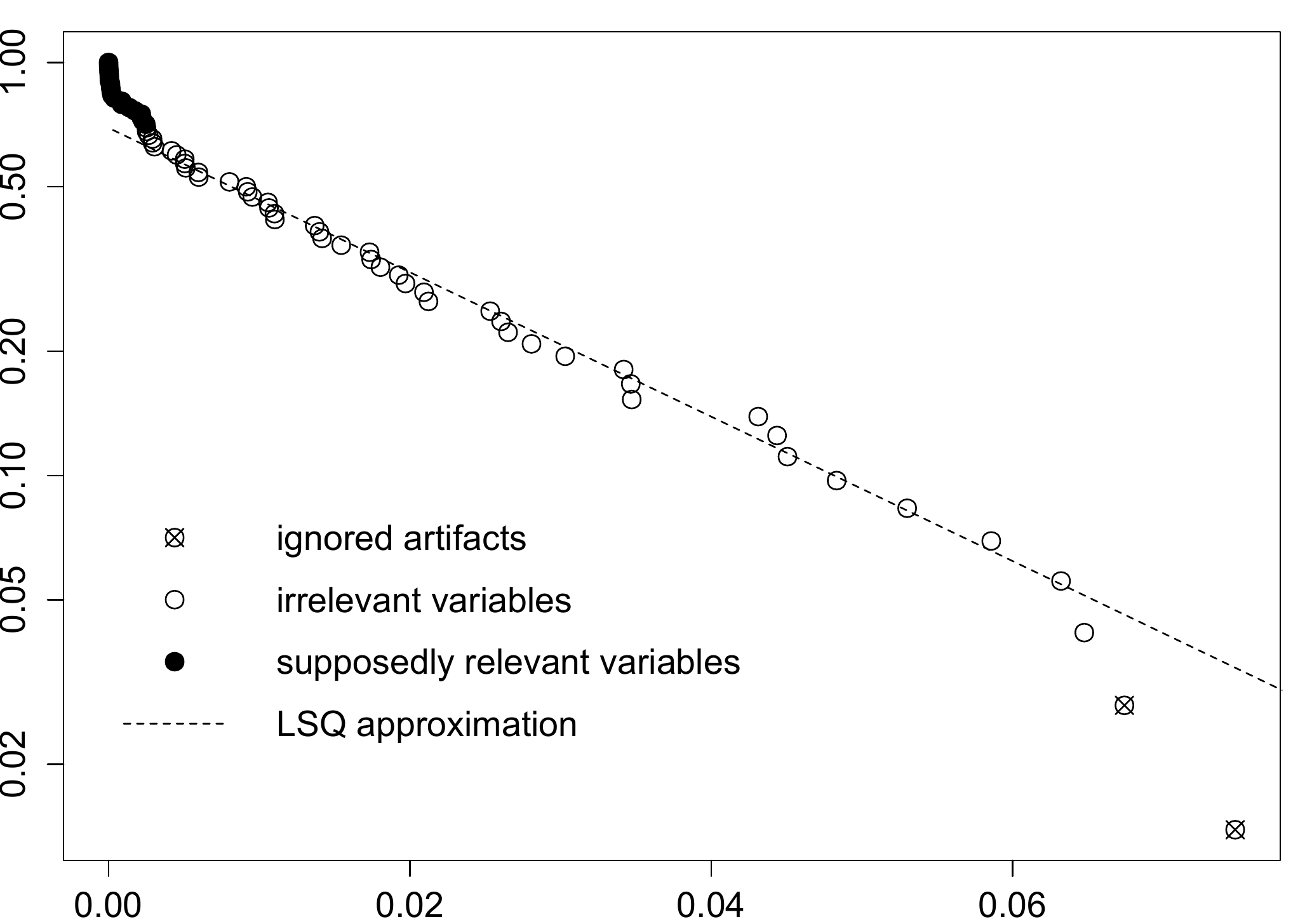}&
\includegraphics[width=0.48\textwidth]{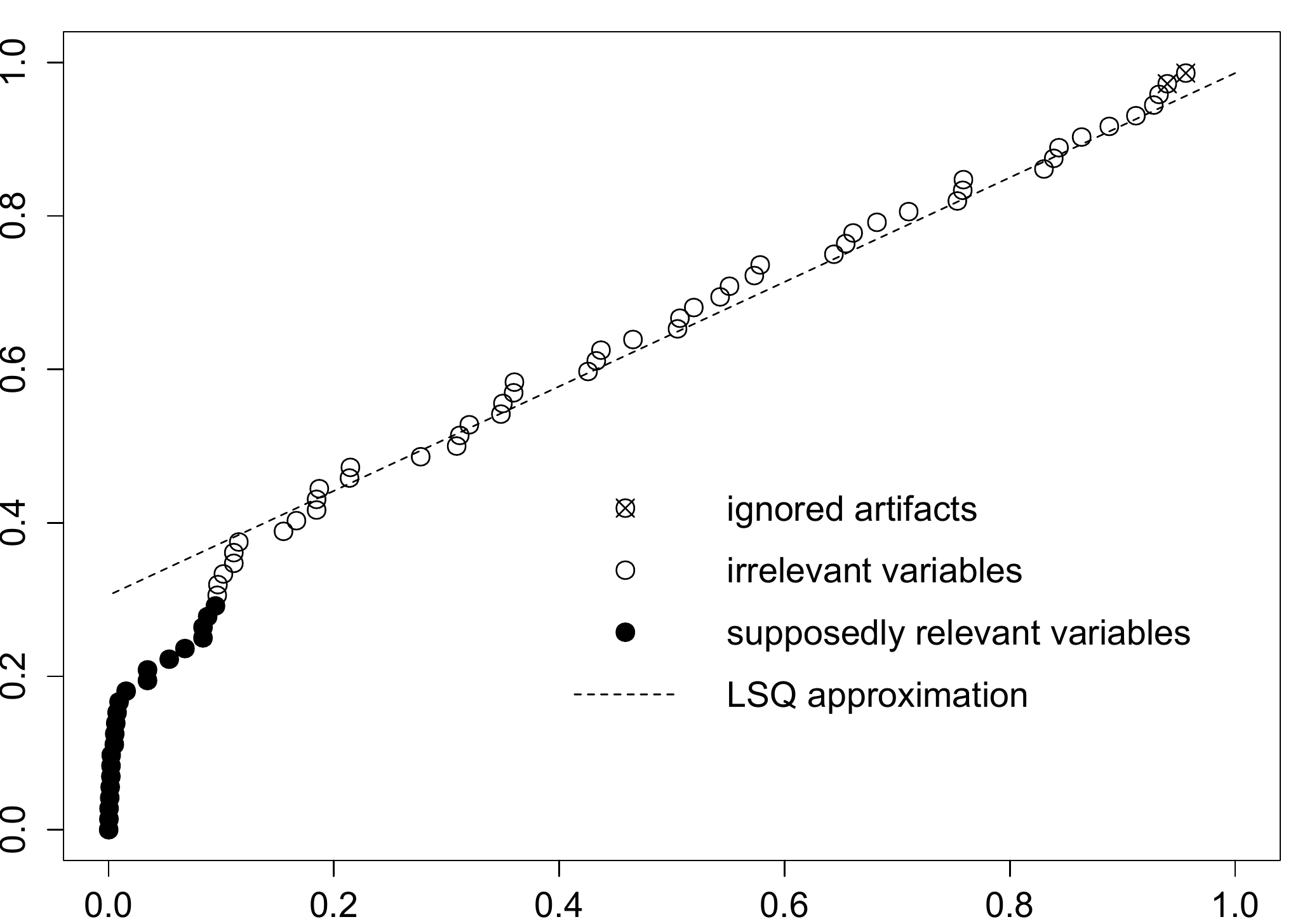}\\
\end{tabular}
\caption{Distribution of $p_{min}$ 
for the dataset used in the  Appendix~\ref{xmp_weaker}.
The set contains 72 variables, 22 of which are relevant, and the remaining ones are random.
a) P-P plot of the minimum $\chi^2$ $p$-value for the variables in logarithmic scale;
b) P-P plot of the experimental vs. exponential distribution with the estimated parameter $\gamma$.
Two outlier points with high p-value were removed to obtain best fit.}
\label{est_gamma}
\end{figure}

One additional precaution is necessary. 
One can expect the average value of $p_{min}$ to be of the order of the reciprocal of the variable number, so even a handful of incorrect values close to 1 can change it dramatically. 
Thus, the procedure described above is vulnerable to numerical artifacts that can produce some pathologically big values of $p_{min}$. 
To avoid that effect and assure numerical stability in the presence of numerical artifacts, the outlying large $p$-values should be ignored. 
The number of ignored points can be chosen to minimize the weighted square error of the estimated coefficient $\gamma$. 
An illustration of the procedure is shown in Fig.~\ref{est_gamma}. 

%% file: appendix_b.tex
\subsection{Correlated synergistic variables}
\label{problem_synergy}
Consider two correlated binary variables $X_1, X_2$, and the response variable $Y=X_1 \land X_2$ i.e. $Y=1$ if both $X_1$ and $X_2=1$, otherwise $Y=0$. 
The correlations between $X_1$ and $X_2$ as well as a decision variable are shown in the following table 
\begin{center}
\begin{tabular}{cc|cc||cc|cc|}
       &   & $X_1$           &               & 			&  	& $X_1$ & 	\\
       &   & $0$             & $1$           & 			& Y	& 0		& 1 \\
 \hline
       &     &               &               & 			& 	&		&  \\[-1em]   
 $X_2$ & $0$ & $\frac{1}{3}$ & $\frac{1}{6}$ & $X_2$ 	& 0 & 	0	& 0 \\[+0.2em] 
       & $1$ & $\frac{1}{6}$ & $\frac{1}{3}$ & 			& 1	&	0	& 1 \\[+0.2em] 
 \hline      
\end{tabular} 
\end{center}

The dependence between $Y$ and $X_1$, $X_2$  is a classical example of the epistasis in genetics, as defined by \citec{Bateson in 1909}{bateson1909}. However, the interaction information $I_{int}(Y,X_1,X_2)=0$ due to the mutual dependence between $X_1$ and $X_2$. 

\subsection{Statistical strength vs. importance of variables}
\label{xmp_weaker}

\begin{figure}[t]
\centering
\includegraphics[width=0.48\textwidth]{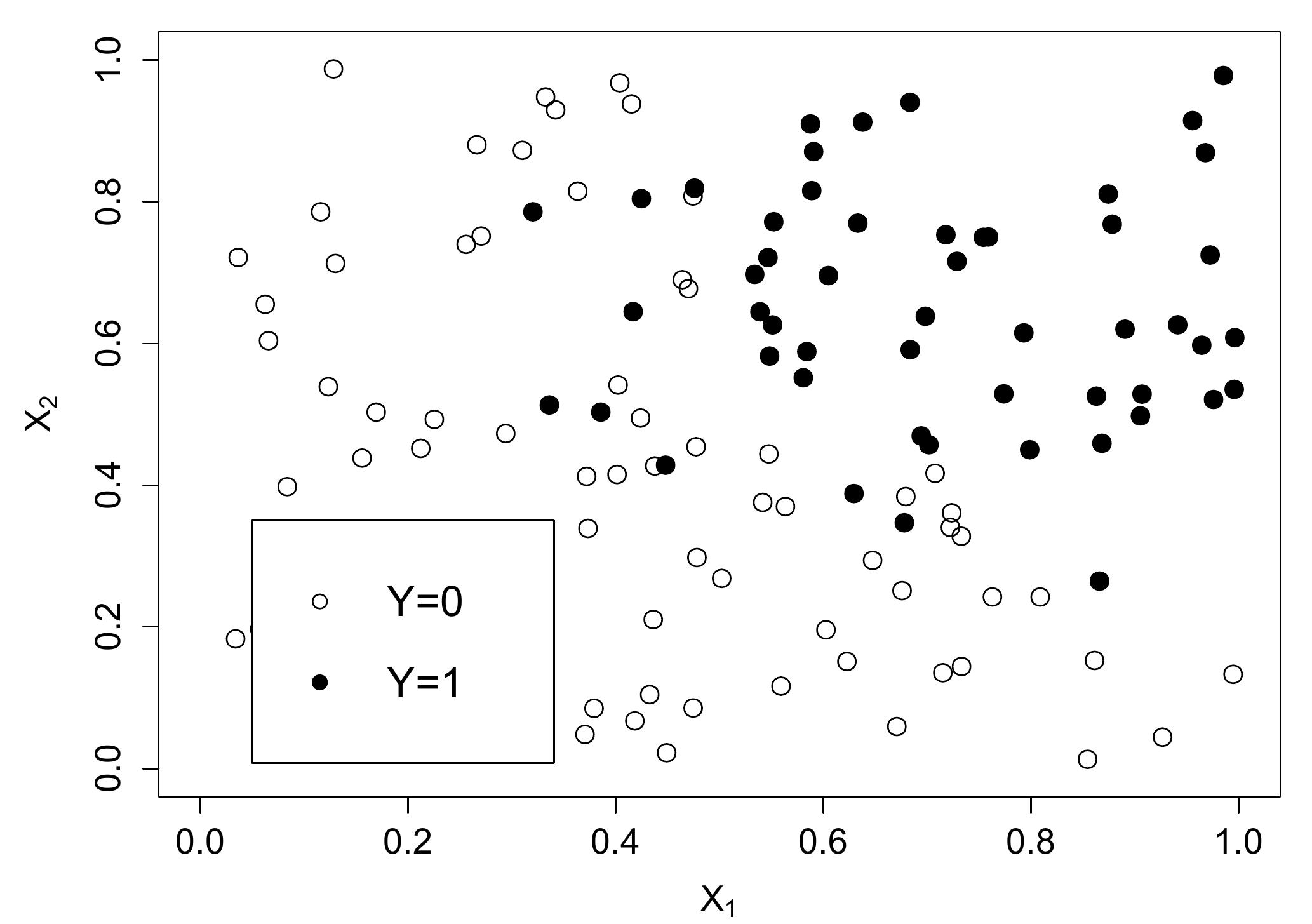}
\includegraphics[width=0.48\textwidth]{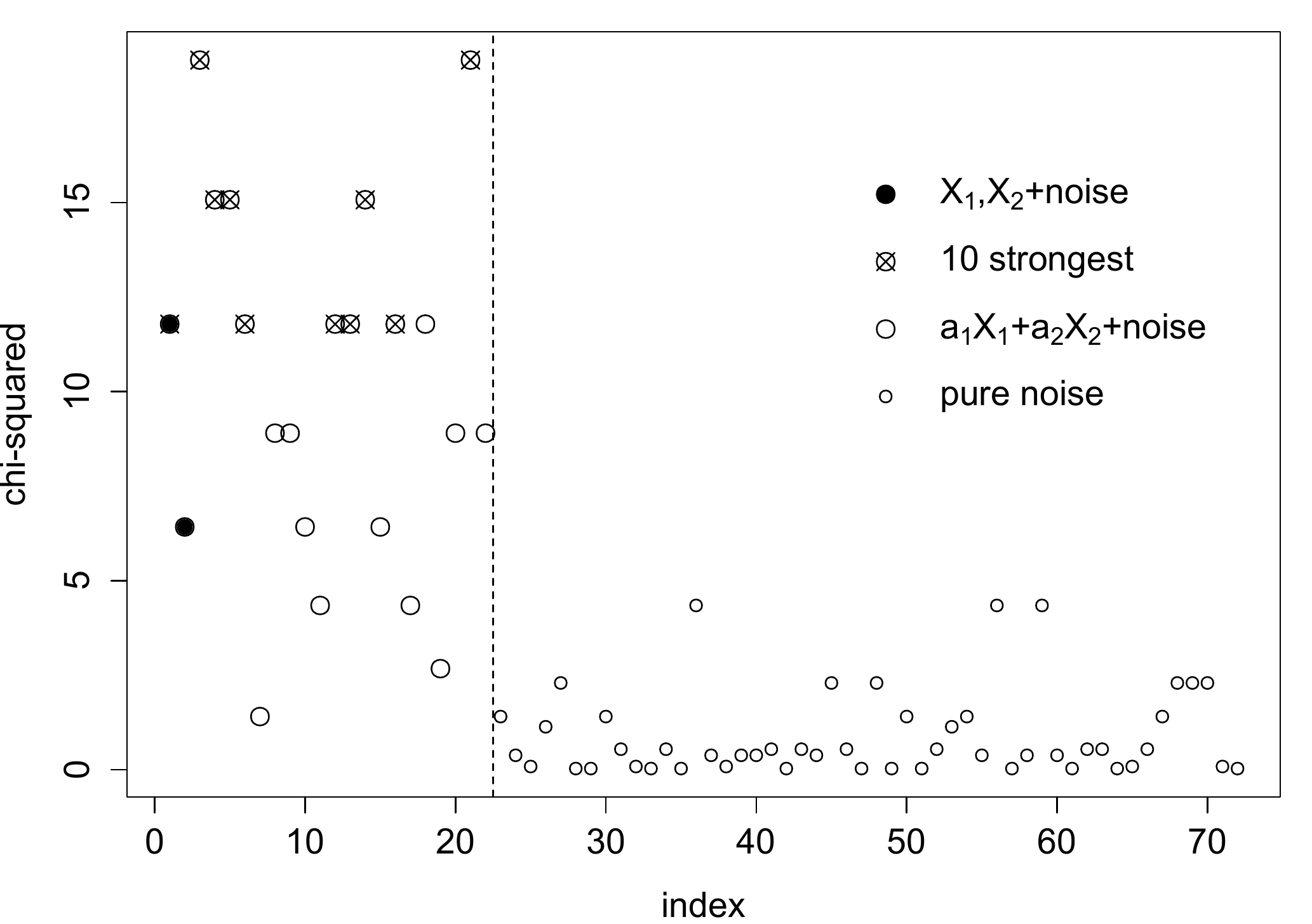}
\caption{The example 2-dimensional dataset and the results of the $\chi^2$ test 
for a noisy dataset. 
}
\label{fig_weaker}
\end{figure}

In the example dataset the response variable $Y$ is a function of two variables $X_1$, $X_2$, defined as: 
\{$Y = 1$ if $X_1 >0.5$ and $X_2 >0.5$; $Y = 0$ otherwise\}. 
The descriptor set consists of $X_1$, $X_2$ and additionally their twenty linear combinations fifty random variables. 
The result of the univariate $\chi^2$ test is the same for all of the 22 relevant variables (result not shown).
When 50\% noise is added to each variable, what simulates the real data (left panel of Fig.~\ref{fig_weaker}), $X_1$ and $X_2$ are no longer the statistically strongest variables.  
Only one of them lies among the 10 strongest ones (right panel of Fig.~\ref{fig_weaker}). 
A procedure that restricts search to a few statistically strongest features could ignore the most important ones. 

\subsection{Multivariate interactions}

The example datasets discussed in this section consist of 400 random samples of variables drawn from the same distribution. 

\subsubsection{Relevant, redundant and nuisance variables}
\label{xmp_multi}
\begin{figure}[t]
\centering
\begin{tabular}{ll}
{\bf a)}&{\bf b)}\\
\includegraphics[width=0.48\textwidth]{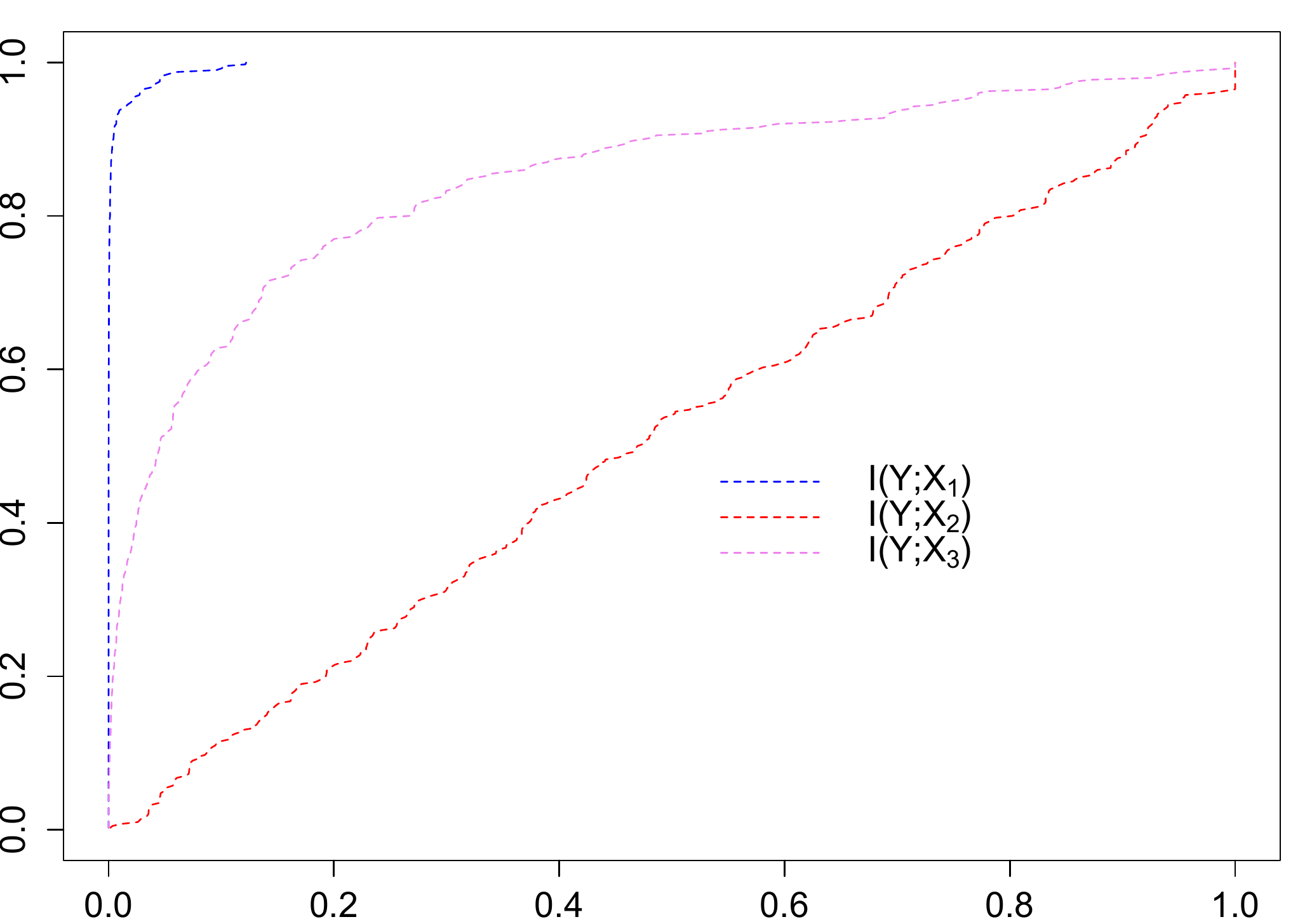}&
\includegraphics[width=0.48\textwidth]{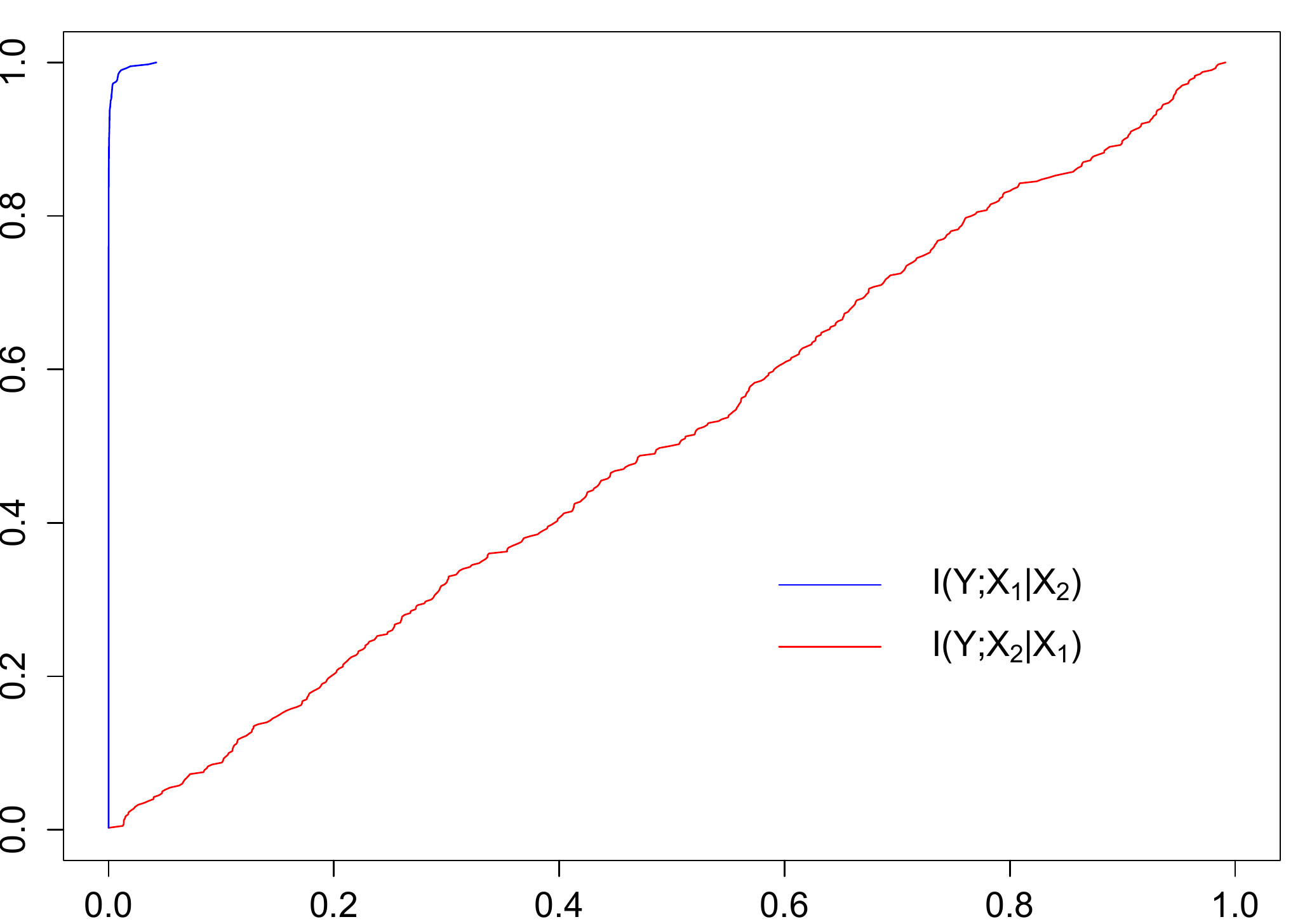}\\
{\bf c)}&{\bf d)}\\
\includegraphics[width=0.48\textwidth]{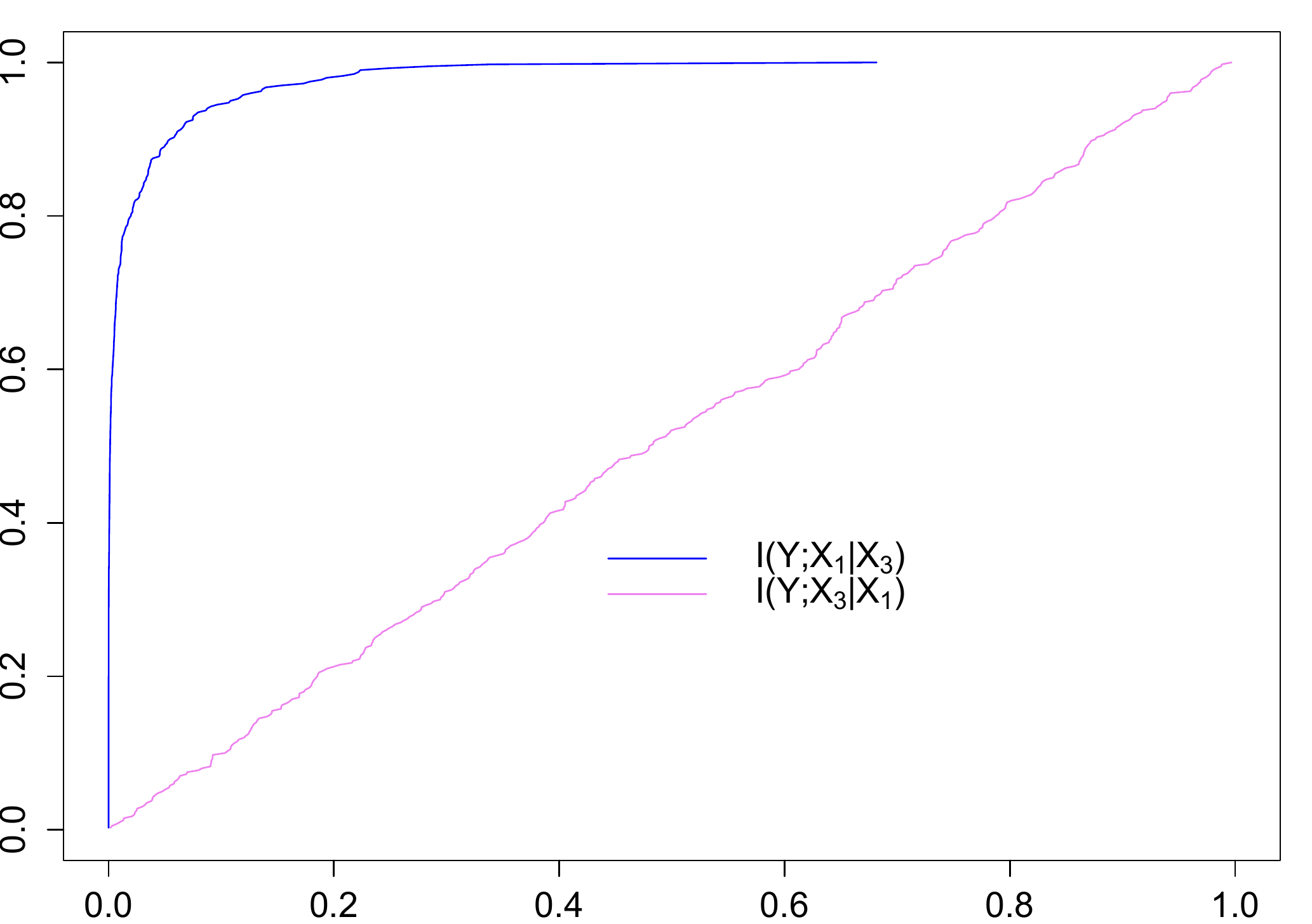}&
\includegraphics[width=0.48\textwidth]{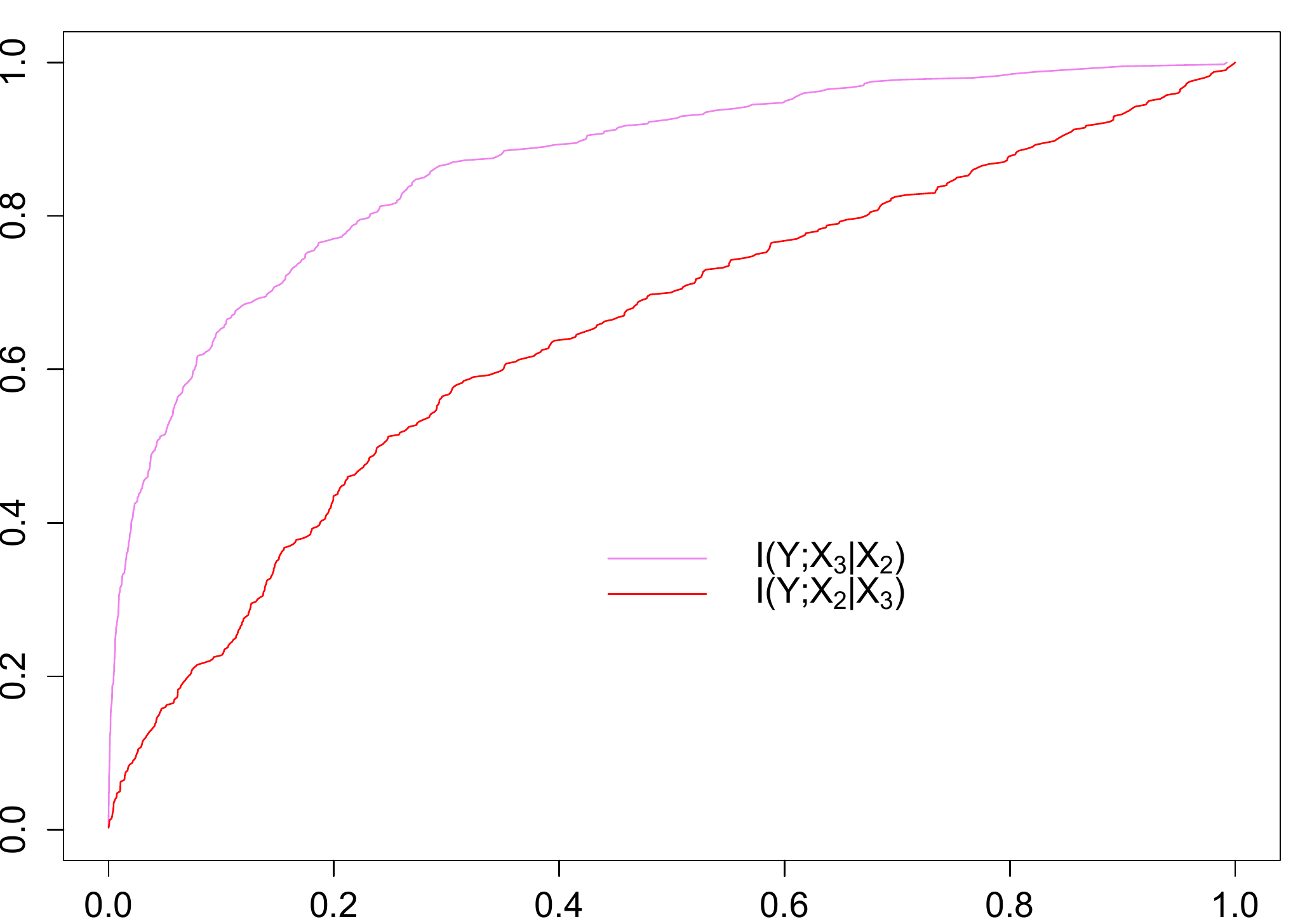}\\
\end{tabular}
\caption{P-P plots of univariate mutual information (\emph{MI}) and conditional mutual information (\emph{CMI}) for the relevant, redundant and nuisance variables: a)\emph{MI} for variables $X_1$, and $X_2$, b) \emph{CMI} for  $(X_1, X_2)$, c) \emph{CMI} for $(X_1,X_1+X_2)$ d) \emph{CMI} for $(X_2,X_1+X_2)$. 
}
\label{xmp_rel}
\end{figure}

The first example ilustrates notions of relevant, redundant and nuisance variables. The variables are defined as follows:
\begin{description}
\item{$X_1$ and $X_2$} are mutually independent random binary variables;
\item{$X_3$} is a mixture of $X_1$ and $X_2$ i.e. $X_3$ is equal to $X_1$ for randomly chosen 50\% of objects and to $X_2$ for the remaining 50\%.
\item {$Y=X_1+random\; noise$} (random noise is added to for clarity of P-P plots -- otherwise the relevant variables would be represented by vertical lines at $0$).
\end{description}
Fig. \ref{xmp_rel} shows P-P plots of the uni-variate mutual information tests for each variable and the conditional mutual information (in both directions) for each pair of variables. 
In this set, the $X_1$ and $X_3$are both informative (top left panel), however, $X_3$ is redundant in pair with $X_1$ (bottom left panel). 
Variable $X_2$ is not informative, neither alone (top left) nor in pair with $X_1$ (top right). 
However, it is informative, when considered in pair with $X_3$ (bottom right). 
The mutual information between $Y$ and the irrelevant variable $X_2$ fits the theoretical $\chi^2$ distribution, unlike those for the relevant $X_1$ and the mixture $X_3$. 

Similarly, when the pair $(X_1,X_3)$ is considered, the variable $X_3$ contains less information about the response variable than $X_1$. 
It is redundant in this subset, so the conditional mutual information $\hat{I}(Y,X_3|X_1)$ follows the $\chi^2$ distribution.
Interestingly, in the pair $(X_3,X_2)$, both $X_3$ and  $X_2$ are informative. 
$X_2$ is an example of a nuisance variable. 
Such variable has no direct association with the response variable, however it modifies the value of the truly relevant variable.  
The so called "lab", "technician" or "house" effects fall into this category. 
While measurement of some variable theoretically should be independent on the person performing the experiment or laboratory where it is performed, nevertheless, such effects happen in practice. 
In such a case, the information on the true variables is not available, only the values modified by the known, or unknown nuisance factors. 
If the pure variable $X_1$ is not available, the knowledge of $X_2$ can be used to refine the signal of $X_3$. 
Although not directly associated with the response variable, such a variable may be useful for model building. 
Such effects may be undetectable by a univariate test.

\subsubsection{Pure synergy of independent variables}
\begin{figure}
\centering
\begin{tabular}{ll}
{\bf a)}&{\bf b)}\\
\includegraphics[width=0.48\textwidth]{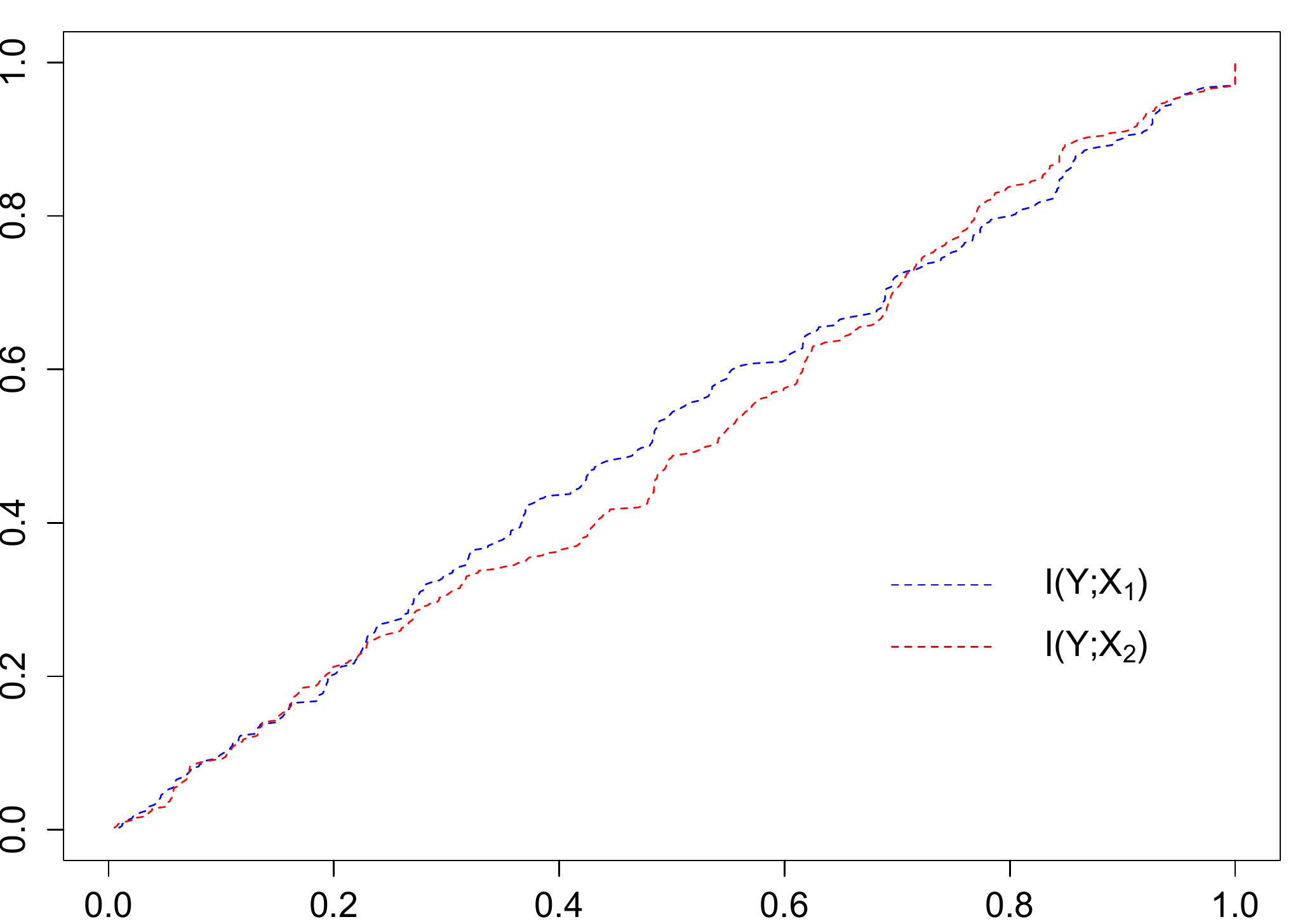}&
\includegraphics[width=0.48\textwidth]{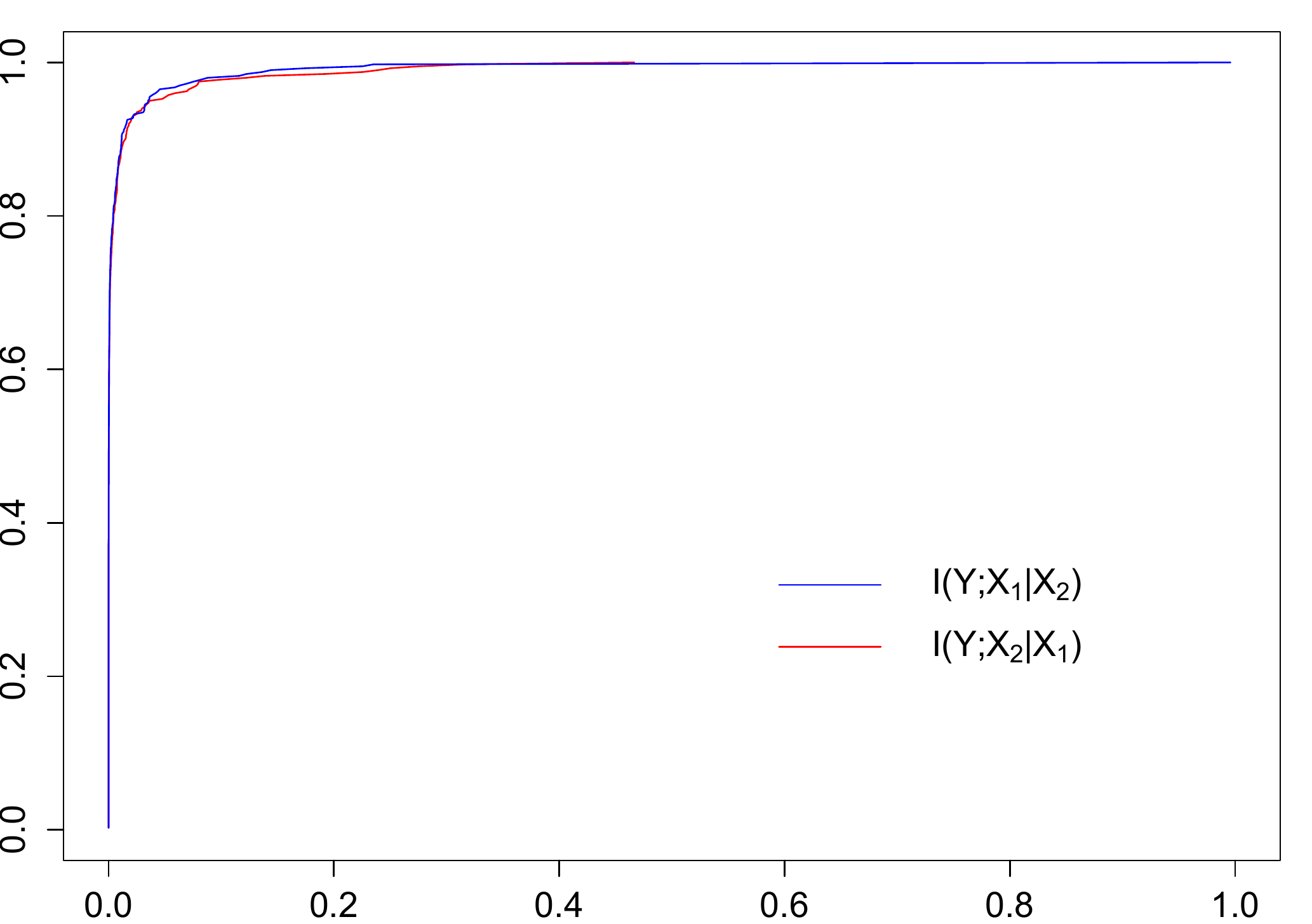}\\
\end{tabular}
\caption{P-P plots of univariate mutual information (\emph{MI}) and conditional mutual information (\emph{CMI}) for the synergistic variables: a) univariate \emph{MI} for variables $X_1$, and $X_2$, b) \emph{CMI} in both directions for pair $(X_1, X_2)$}
\label{xmp_synr}
\end{figure}

In this example, the variables are defined as follows:
\begin{description}
\item{$X_1$ and $X_2$} are mutually independent random binary variables;
\item{$Y=(X_1=X_2)+random\; noise$} i.e. $Y=1$ if $X_1$ and $X_2$ are equal to one another, otherwise $Y=0$. 
\end{description}
The test results are presented in Fig.~\ref{xmp_synr}. The univariate test shows that none of the variables proves relevant alone. However, the test of conditional mutual information for the pair of variables discovers the relevance of both $X_1$ and $X_2$. This is a classic example of synergy.

\subsubsection{Epistasis of correlated variables}

\begin{figure}
\centering
\begin{tabular}{ll}
{\bf a)}&{\bf b)}\\
\includegraphics[width=0.48\textwidth]{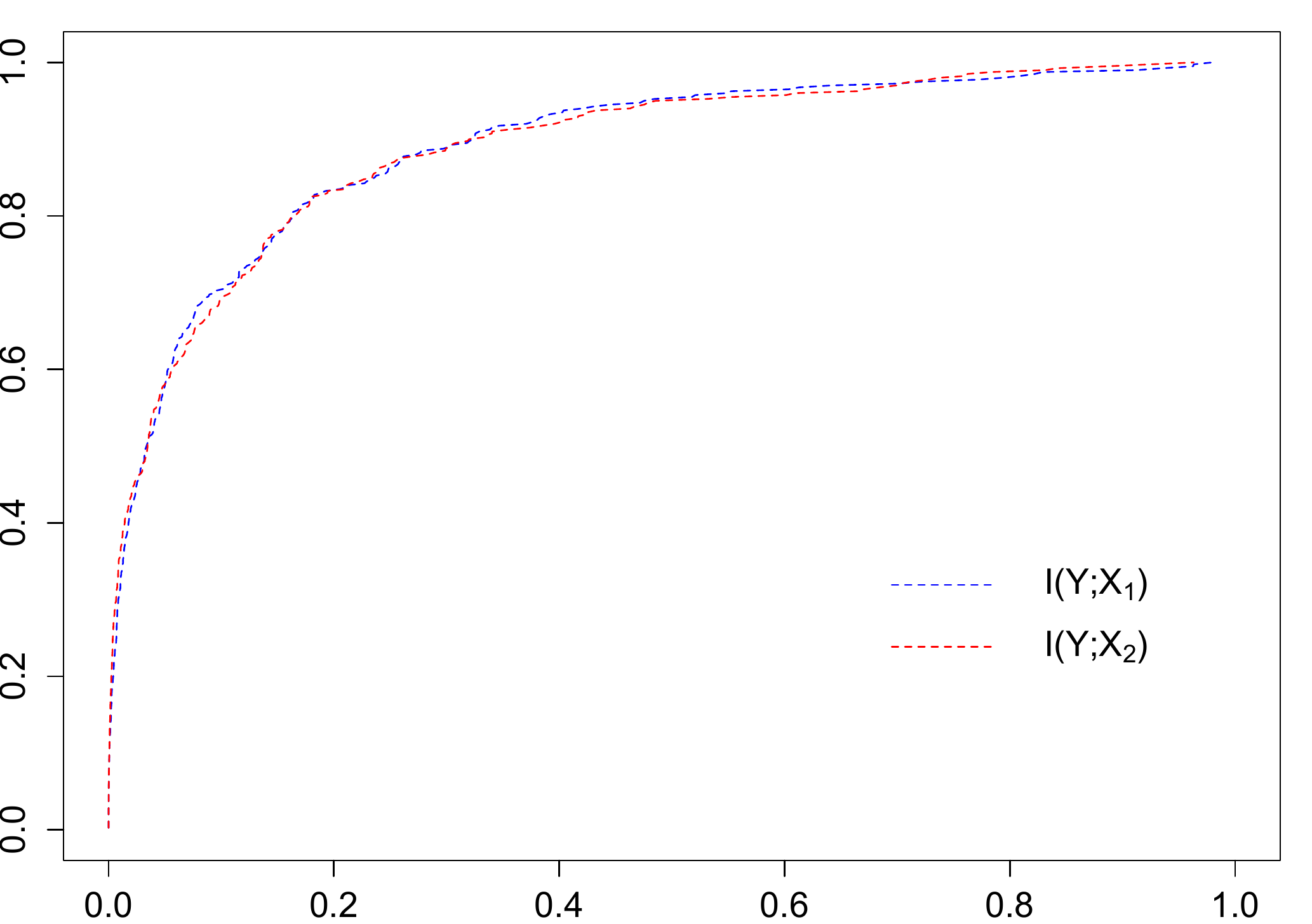}&
\includegraphics[width=0.48\textwidth]{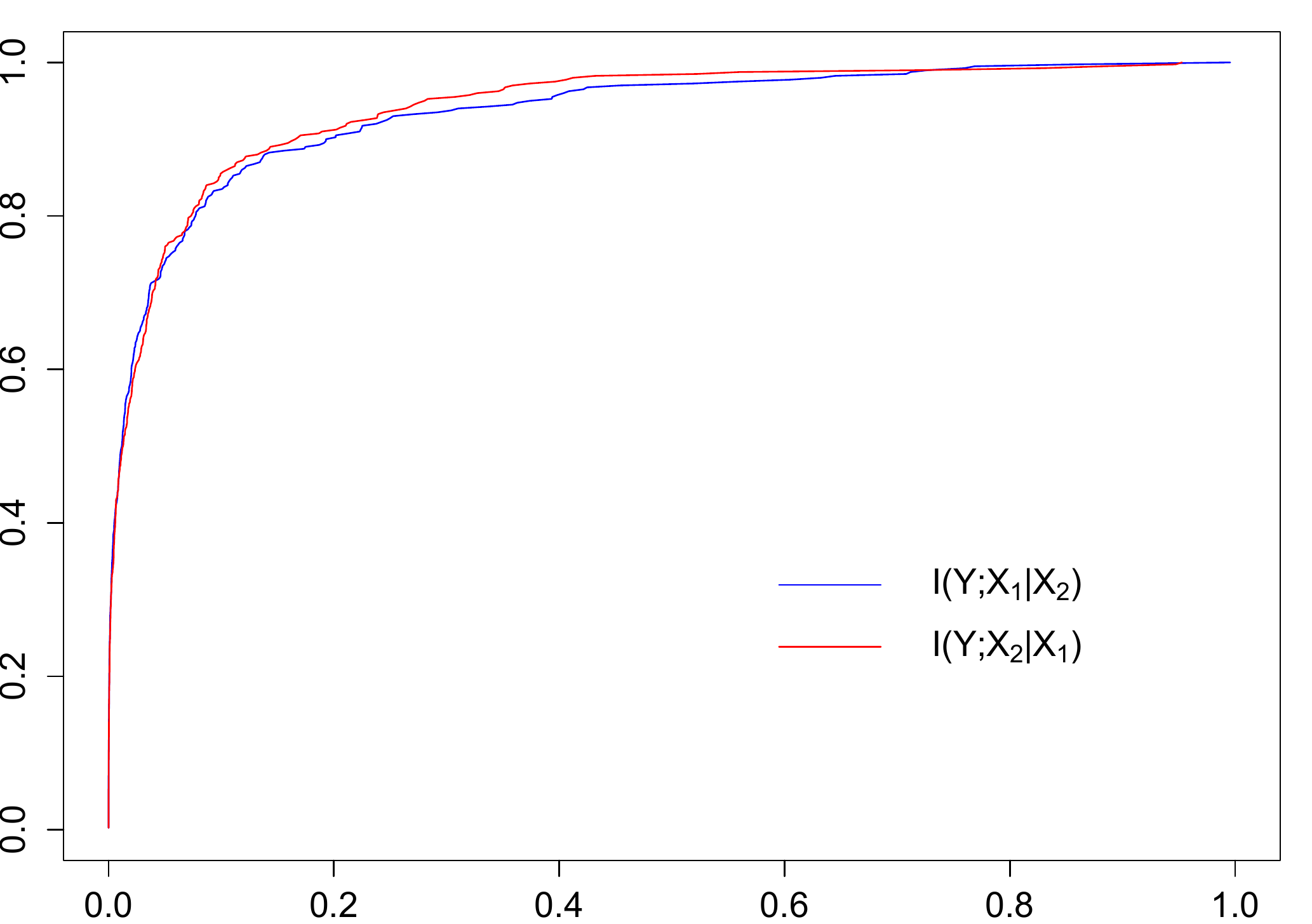}\\
\end{tabular}
\caption{P-P plots of univariate mutual information (\emph{MI}) and conditional mutual information (\emph{CMI}) for the correlated variables, involved in epistatic interaction: a) univariate \emph{MI} for variables $X_1$, and $X_2$, b) \emph{CMI} in both directions for pair $(X_1, X_2)$}
\label{cor_epi}
\end{figure}

This example explores the problem described in Appendix~\ref{problem_synergy}. 
\begin{description}
\item{$X_1$ and $X_2$} are correlated with one another;
\item{$Y=(X_1\land X_2)+random\; noise$} i.e. $Y=1$ if both $X_1$ and $X_2$ are equal 1.
\end{description}
The relevance of the variables $X_1$, $X_2$ can be detected by the univariate test, and the result of the conditional mutual information test is even stronger (see Fig.~\ref{cor_epi}).
